\pdfoutput=1

\documentclass[11pt]{article}

\usepackage[final]{acl}

\usepackage{times}
\usepackage{latexsym}

\usepackage[T1]{fontenc}

\usepackage[utf8]{inputenc}

\usepackage{microtype}

\usepackage{inconsolata}

\usepackage{graphicx}

\usepackage{enumitem}
\usepackage{booktabs}
\usepackage{multirow}
\usepackage{geometry}
\usepackage{amsmath}
\usepackage[most]{tcolorbox}
\usepackage{amssymb}
\usepackage{wrapfig}
\usepackage{lipsum}
\usepackage{array}
\usepackage{colortbl}
\usepackage{graphicx}
\usepackage{subcaption}
\usepackage{makecell}
\usepackage{soul} 
\usepackage{marginnote}
\usepackage{caption}
\usepackage{float}

\newtcolorbox{boxK}{
    sharpish corners, 
    boxrule = 0pt,
    toprule = 4.5pt, 
    enhanced,
    fontupper = \small,
    fuzzy shadow = {0pt}{-2pt}{-0.5pt}{0.5pt}{black!35} 
}

\definecolor{reflact}{RGB}{233,242,255}
\definecolor{reflection}{RGB}{0,92,211}
\definecolor{nothinking}{RGB}{102,106,115}
\definecolor{thought}{RGB}{235,131,71}
\definecolor{obs}{RGB}{150, 82, 214}
\definecolor{action}{RGB}{0, 114, 6}
\definecolor{custom_green}{RGB}{1, 171, 79}

\title{\raisebox{-0.1cm}{\includegraphics[width=0.95cm]{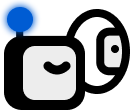}} ReflAct: World-Grounded Decision Making in\\LLM Agents via Goal-State Reflection}
 
\author{
 \textbf{Jeonghye Kim\textsuperscript{1$\ast$}},
 \textbf{Sojeong Rhee\textsuperscript{1$\ast$}},
 \textbf{Minbeom Kim\textsuperscript{2}},
 \textbf{Dohyung Kim\textsuperscript{2}},
\\
 \textbf{Sangmook Lee\textsuperscript{2}},
 \textbf{Youngchul Sung\textsuperscript{1$\dagger$}},
 \textbf{Kyomin Jung\textsuperscript{2$\dagger$}},
\\
\\
 \textsuperscript{1}KAIST,
 \textsuperscript{2}Seoul National University
\\
 \small{
   \{jeonghye.kim, sojeong.rhee, ycsung\}@kaist.ac.kr, kjung@snu.ac.kr
 }
}

\begin{document}
\maketitle

\footnotetext{\textsuperscript{$\ast$} Equal contribution.}
\footnotetext{\textsuperscript{$\dagger$} Co-corresponding authors.}

\begin{abstract}

Recent advances in LLM agents have largely built on reasoning backbones like ReAct~\citep{react}, which interleaves thought and action in complex environments. However, ReAct often produces ungrounded or incoherent reasoning steps, leading to misalignment between the agent’s actual state and goal. Our analysis finds that this stems from ReAct’s inability to maintain consistent internal beliefs and goal alignment, causing compounding errors and hallucinations. To address this, we introduce ReflAct, a novel backbone that shifts reasoning from merely planning next actions to continuously \textit{reflecting on the agent’s state relative to its goal}. By explicitly grounding decisions in states and enforcing ongoing goal alignment, ReflAct dramatically improves strategic reliability. This design delivers substantial empirical gains: ReflAct surpasses ReAct by 27.7\% on average, achieving a 93.3\% success rate in ALFWorld. Notably, ReflAct even outperforms ReAct with added enhancement modules (e.g., Reflexion, WKM), showing that strengthening the core reasoning backbone is key to reliable agent performance.

\end{abstract}

\section{Introduction}

Recent advancements in Large Language Models (LLMs) have significantly enhanced their reasoning capabilities, enabling LLM-based agents to perform complex multi-step decision making beyond static problem solving \cite{mathsurvey, AlphaGeometry2}. As LLMs are extended into agentic frameworks where they interact with open-ended environments, the ability to solve long-horizon tasks through ongoing interaction and sequential reasoning has become a central research focus \cite{survey1, eto}.

A large number of recent LLM agents build on the ReAct~\cite{react} framework, which integrates Chain-of-Thought (CoT) reasoning~\cite{cot} with an action selection mechanism. Various extensions have been proposed to enhance this backbone by incorporating reflective iterations or memory components to improve reasoning performance~\cite{reflexion, rafa, wkm}. However, mounting evidence suggests that  ReAct and its variants often deliver limited benefits and can even deteriorate performance, especially in complex, partially observable, or dynamic environments~\cite{NoThinking, agentboard, brittlereact}. These findings raise questions about the current direction of extending the basic ReAct reasoning-action framework.

\begin{figure}[t!]
\centering
  \includegraphics[width=\linewidth]{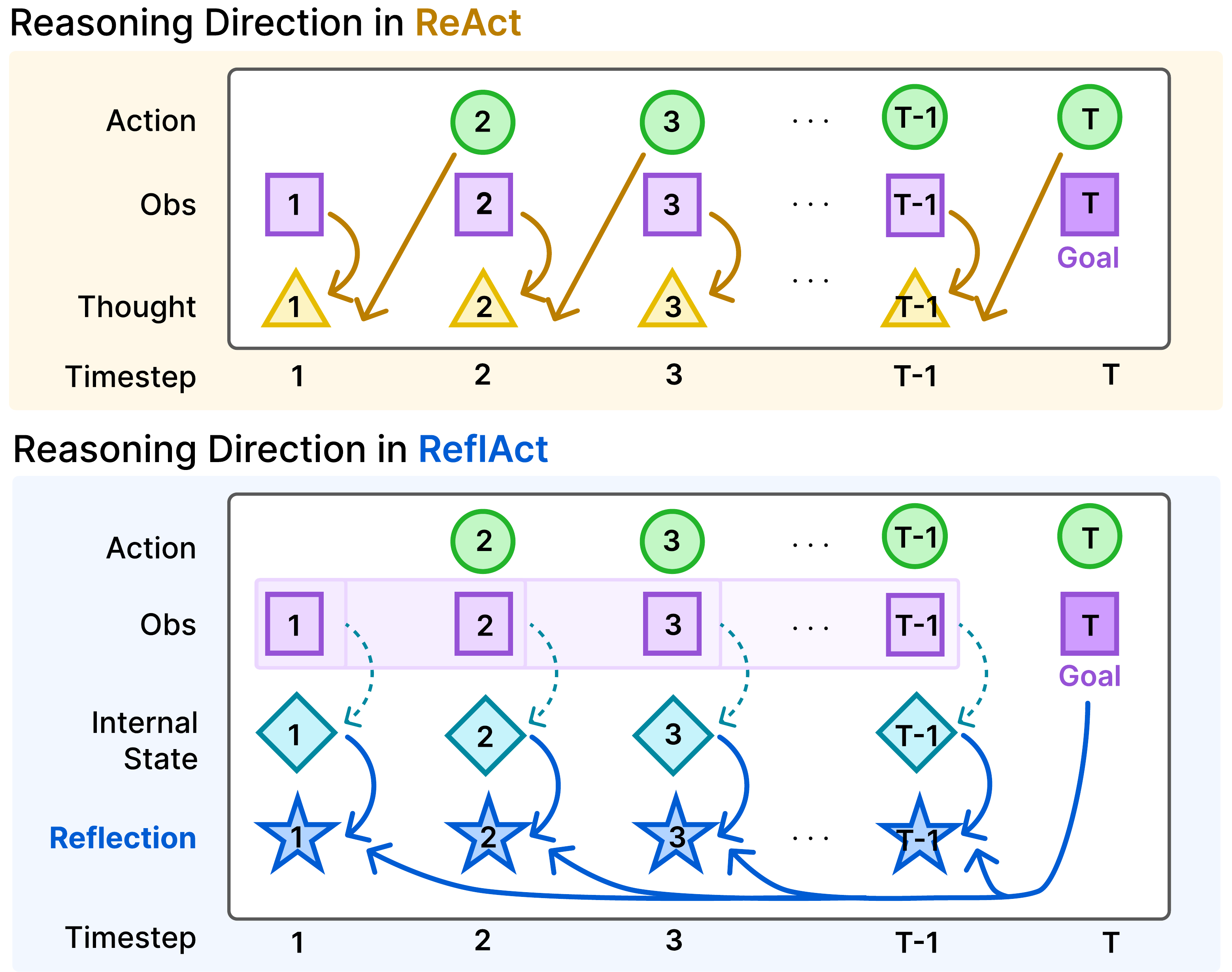}
  \caption{Comparison of reasoning influence in ReflAct and ReAct. While ReAct focuses on the current observation and the next action at each timestep, ReflAct reflects on the internal belief state and the task goal.}
  \label{fig:direction_of_thought}
  \vspace{-6mm}
\end{figure}

In this paper, we claim that true progress lies in fundamentally redesigning the backbone reasoning process itself, rather than adding complementary modules.
Our analysis shows that the fundamental problem of current ReAct-based reasoning methods is the frequent lack of proper grounding of generated thoughts on the agent's historical context or overarching objective. These misaligned thoughts guide subsequent action choices. In dynamic or partially observable environments, such reasoning errors compound over time, leading to increasingly divergent internal beliefs, deteriorated decision quality, and failed task execution.

To address this limitation, we introduce \textbf{ReflAct (Reflect for Action)}, a new backbone framework that shifts the focus of the agent's thought from \textit{predicting the next action} to continuously \textit{reflecting on its current state in the context of the task goal}, as shown in Figure \ref{fig:direction_of_thought}. Unlike previous approaches that layer reflective or memory modules onto the ReAct backbone \cite{reflexion, wkm, mpo}, ReflAct does not rely on such additional components but replaces the core reasoning–action cycle itself. That is, at each timestep, the agent evaluates whether its current trajectory and situation align with its intended long-term objective before selecting an action. This reflective thought mechanism enables early detection of potential deviations and facilitates timely strategy adjustments, as we shall see shortly. By grounding decisions in actual observations and maintaining continuous goal alignment, ReflAct substantially reduces hallucination tendency and enhances long-term strategic coherence.

We implemented ReflAct using both open-source (Llama-3.1-8B/70B-Instruct \cite{llama3}) and proprietary (GPT-4o-mini/4o \cite{gpt-4o}) LLMs, and evaluated it in three text-based environments: ALFWorld \cite{alfworld}, ScienceWorld \cite{scienceworld} and Jericho \cite{jericho}, which require an agent to decompose the goal and acts over a long time horizon under partial observability. Empirical results show that ReflAct helps the agent better align its internal beliefs with the environment and promotes goal-consistent behavior. As a result, ReflAct significantly outperforms ReAct by 36.4\%, 8.5\%, and 38.1\% on ALFWorld, ScienceWorld, and Jericho, respectively. Notably, ReflAct also surpasses enhancement modules layered on top of existing backbones, demonstrating that revising the reasoning process itself can be more effective than adding new components.

\section{Preliminaries}

We model the agent's task in a language-based environment as a Partially Observable Markov Decision Process (POMDP) \cite{puterman1990markov}, defined as $\mathcal{M} = \left<\mathcal{U}, \mathcal{S}, \mathcal{A}, \mathcal{O}, \mathcal{P}, \mathcal{R} \right>$, where $u \in \mathcal{U}$ is a task instruction, $s \in \mathcal{S}$ is the (hidden) environment state, $a \in \mathcal{A}$ is an action, $o \in \mathcal{O}$ is an observation, $\mathcal{P}$ is the transition function, and $\mathcal{R}$ is the reward function. In our setting, $\mathcal{U}$, $\mathcal{A}$ and $\mathcal{O}$ are expressed in natural language.

\noindent \textbf{The ReAct Framework} ~~ The ReAct framework \cite{react} enables LLM agents to solve tasks through interleaved reasoning and action. To incorporate reasoning steps (thoughts), we extend the above POMDP  by including a thought space $\mathcal{T}$, where each thought $\tau \in \mathcal{T}$ represents a natural language reasoning step generated by the agent. 


Then, this extended POMDP operates as follows. Given a task instruction $u \in \mathcal{U}$, the agent follows a policy $\pi_\theta$ that alternates between generating a thought as $\pi_\theta^{\text{thought}}$ and taking an executable action as $\pi_\theta^{\text{act}}$. At each time step $t$, the agent receives an observation $o_t$ and forms a context $c_t = (h_t, o_t)$, where $h_t =  \{u, \tau_1, a_1, o_1, \dots, \tau_{t-1}, a_{t-1}\}$ is the history of previous interactions. Based on $c_t$, the agent samples a thought from a context-conditioned distribution: $
\tau_t \sim \pi_\theta^{\text{thought}}(\cdot \mid c_t).
$
This thought $\tau_t$ is then appended to the context to form an enriched input $c'_t = c_t \oplus \tau_t$, which is used to select the next action: $
a_t \sim \pi_\theta^{\text{act}}(\cdot \mid c'_t)$. 
The selected action $a_t$ is executed to the environment, resulting in the next observation $o_{t+1}$. This reasoning–acting loop continues until the task is completed or a predefined time limit is reached.

\section{Revisiting the Efficacy of Thought} \label{sec: motivation}

\subsection{How Thought Impacts Action Selection}

Basically the thought $\tau_t$ functions as a reweighting mechanism for the action probability distribution of the policy $\pi_\theta^{\text{act}}$  by enriching the context $c_t$ into an augmented form: $c'_t = c_t \oplus \tau_t$.

\begin{figure}[h!]
\centering
  \includegraphics[width=0.5\textwidth]{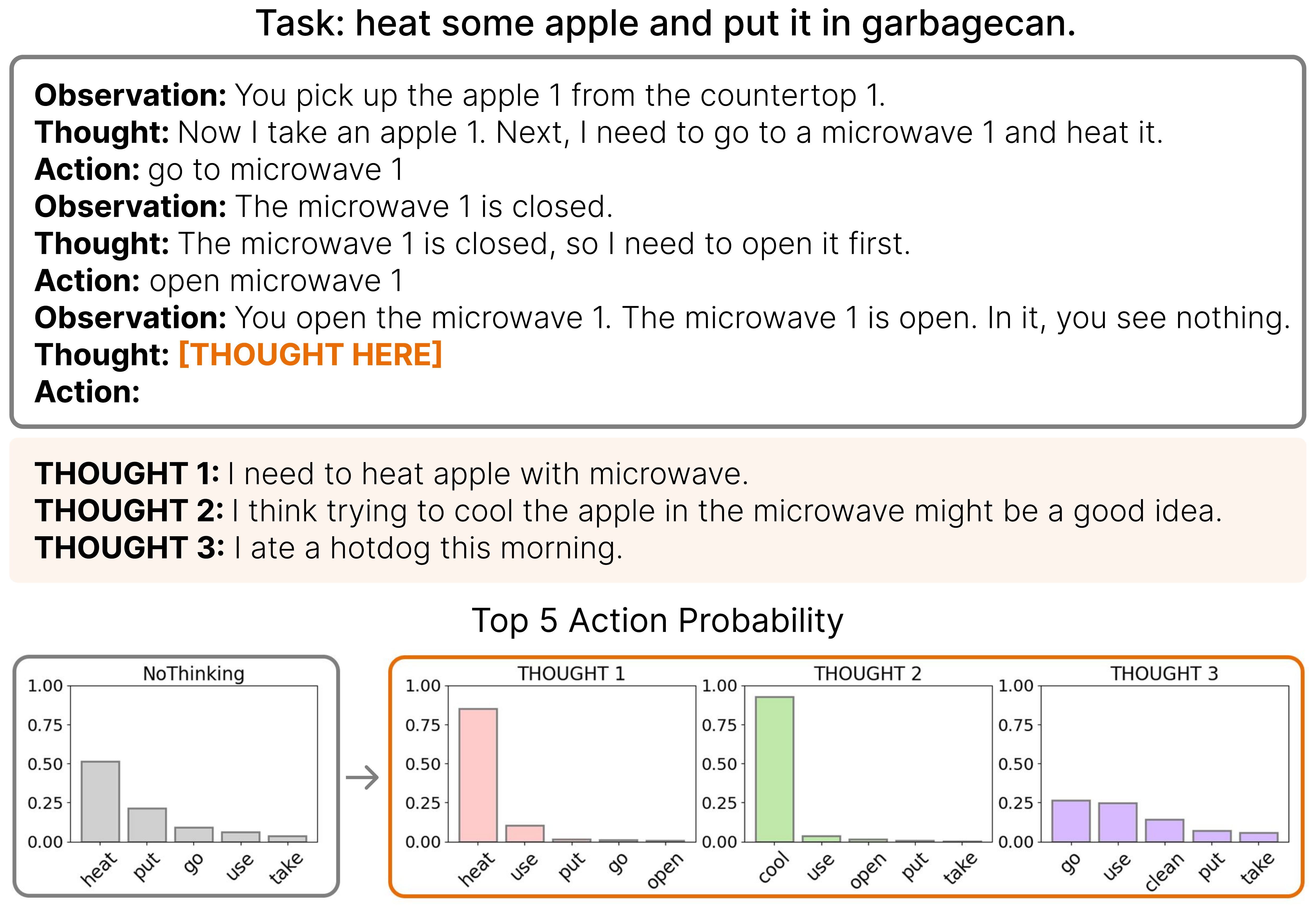}
  \caption{Changes in the action probability distribution when modifying thought in the ALFWorld `heat some apple and put it in garbagecan' task.}
  \label{fig:thought_prob}
\vspace{-3mm}
\end{figure}

We analyzed this process in detail using a task from ALFWorld~\cite{alfworld}, where the task is to find an apple, heat it, and place it in a garbage can.  We implemented the policy $\pi_\theta$ using Llama-3.1-8B-Instruct~\cite{llama3} and compared two agent variants: the NoThinking agent, which selects actions without intermediate reasoning, and the ReAct agent, which generates a reasoning step before each action.  Figure~\ref{fig:thought_prob} shows one instance of execution sequence. The sequence starts with the task description with initial observation, followed by two cycles of thought-action-observation, ending with observation "You open the microwave 1. The microwave 1 is open. $\cdots$". Then, in the next thought step, we tried each of four cases: empty (i.e., no thought) and three thoughts in the middle of Figure~\ref{fig:thought_prob}, and observed the distribution on actions in the next step.

First, without any thought (NoThinking), the agent assigns high probability to the heat action,  still considering other actions such as `put' and `go' with non-negligible probabilities. Now, when a non-empty  thought is provided, the next action probability distribution is directly influenced: \textsc{Thought 1} relevant to the task sharpens the distribution toward appropriate actions, leading to  contextually aligned selections. In contrast, \textsc{Thought 2} and \textsc{Thought 3}, injecting incorrect or conflicting information,  bias the distribution and increase the likelihood of inappropriate actions. 
As seen, thought directly shapes the next action distribution,  either reinforcing correct decisions or causing misalignment depending on its quality. {\em When the model is provided with a structured thought, it relies predominantly on that thought to drive decision making, thereby significantly reducing its dependence on the broader historical context.}

To quantify the overall impact of thought on the agent’s decision-making process, we measured the entropy of the agent’s action probability distribution across 134 tasks from ALFWorld. Specifically, for a given timestep $t$, the entropy is defined as
$$
\mathcal{H}_t^{\text{NoThinking}} = - \sum_{a \in \mathcal{A}} \pi_\theta^{\text{act}}(a \mid c_t) \log \pi_\theta^{\text{act}}(a \mid c_t)
$$
$$
\mathcal{H}_t^{\text{ReAct}} = - \sum_{a \in \mathcal{A}} \pi_\theta^{\text{act}}(a \mid c'_t) \log \pi_\theta^{\text{act}}(a \mid c'_t),
$$
where $c'_t = c_t \oplus \tau_t$ and $\tau_t \sim \pi_\theta^{\text{thought}}(\cdot \mid c_t)$. 
\newcolumntype{C}[1]{>{\centering\arraybackslash}m{#1}}
\begin{table}[h!]
\centering
\small
\renewcommand{\arraystretch}{1.5}
\caption{Average entropy of the action probability distribution when performing 134 ALFWorld tasks using the Llama-3.1-8B-Instruct with NoThinking and ReAct.}
\begin{tabular}{p{3.5cm}|C{1.3cm}C{1.3cm}}
    \hline
    \textbf{Model}  & \textbf{$\bar{\mathcal{H}}^{\text{NoThinking}}$} & \textbf{$\bar{\mathcal{H}}^{\text{ReAct}}$} \\
     \hline
     \hline
    Llama-3.1-8B-Instruct & 1.23 & 0.30 \\
    \hline
\end{tabular}
\label{table:entropy111}
\vspace{-3mm}
\end{table}

We computed the entropy averaged over all timesteps and tasks, and the result is  reported in Table~\ref{table:entropy111}. As shown in the table, the NoThinking agent exhibits  mean entropy of 1.23, whereas the ReAct agent yields a significantly lower value of 0.30. This substantial reduction in entropy implies that conditioning the policy on thought leads to more confident and focused behavior.

\subsection{The Problem of Ungrounded Thought} \label{sec: ungrounded_thought_problem}
Since action $a_t$ is heavily influenced by thought $\tau_t \sim \pi_\theta^{\text{thought}}(\cdot \mid c_t)$, it is crucial that $\tau_t$ accurately captures the core decision-relevant information from the context $c_t = (h_t, o_t)$. If the thought becomes ungrounded, meaning it is disconnected from the actual state, it can mislead the policy and  induce a suboptimal or incorrect action. 

\begin{figure}[h!]
\centering
  \includegraphics[width=0.48\textwidth]{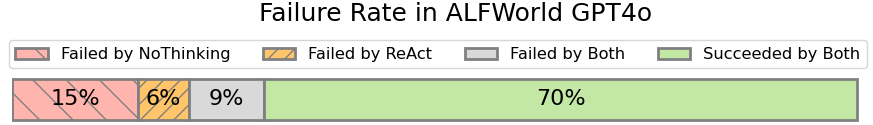}
  \includegraphics[width=0.48\textwidth]{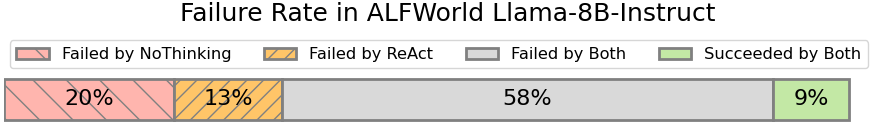}
  \caption{Comparison of failure rates between NoThinking and ReAct on GPT-4o and Llama-3.1-8B-Instruct}
  \label{fig:fail_plot}
  \vspace{-2mm}
\end{figure}

To assess the overall performance degradation from ungrounded thought, we evaluated task failure rates across 134 ALFWorld~\cite{alfworld} tasks using $\pi_\theta$ realized with GPT-4o~\cite{gpt-4o} and Llama-3.1-8B-Instruct~\cite{llama3}. Figure~\ref{fig:fail_plot} categorizes the outcomes into four groups: (1) only the NoThinking agent failed, (2) only the ReAct agent failed, (3) both failed, and (4) both succeeded. As shown in the figure, incorporating thought can help the agent solve tasks that would otherwise fail without it. However, thought also introduces new risks, causing failures that could have been avoided without thought, even in stronger models like GPT-4o.

\begin{figure*}[h!]
\centering
  \includegraphics[width=0.99\textwidth]{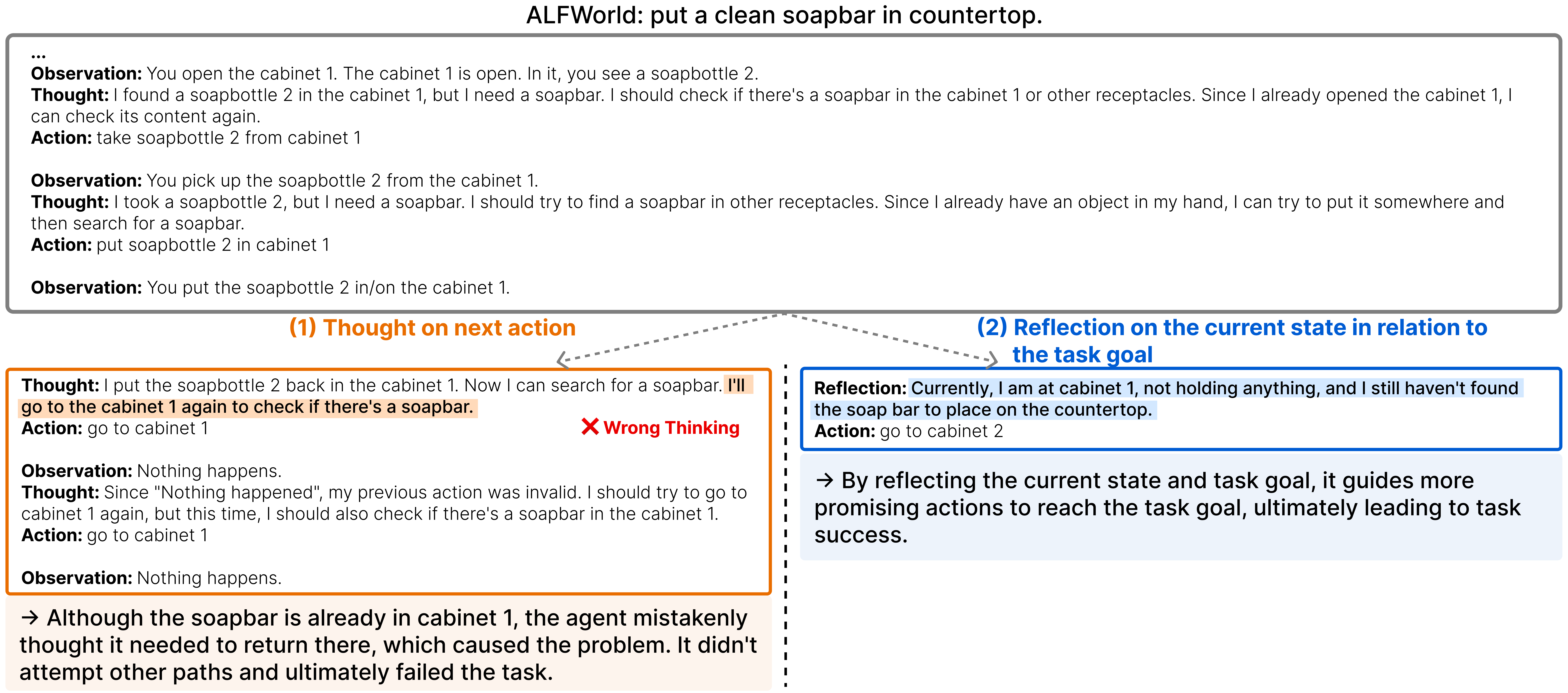}
  \caption{In the ALFWorld task `put a clean soapbar in countertop', the agent picks up a soapbottle from cabinet 1 and puts it back, \textbf{(1)} while planning the next action, it redundantly tries to go back to cabinet 1 and fails, \textbf{(2)} after reflecting on the state and goal, it moves elsewhere and succeeds.}
  \label{fig:when_thoughts_hinder_alfworld}
  \vspace{-0.3cm}
\end{figure*}

To see the cause of the ReAct agent's failure, we examined a failure case of the ReAct agent in ALFWorld~\cite{alfworld}.
As seen in Figure~\ref{fig:when_thoughts_hinder_alfworld} (1), the agent is given a simple task: Find a soap bar, take it and place it on the countertop. After interacting with a soap bottle 2 in cabinet 1, the agent attempts to return to cabinet 1 to search for the soap bar even though it is not there, because the generated thought does not capture the current situation in the context of goal properly from its observation history. This redundant and illogical behavior leads to a loop, and the agent ultimately fails to complete the task. Failure cases in ScienceWorld is provided in Figure~\ref{fig:when_thoughts_hinder_scienceworld} in Appendix~\ref{appendix: another_thought_problem}.

From the examples, we see that  the agent fails due to two major reasoning limitations:
\begin{enumerate}[label=(\arabic*), itemsep=0pt, topsep=2pt, leftmargin=*]
   \item \textbf{A lack of grounding in its internal state.} The agent fails to maintain a coherent internal state, leading to inconsistency, e.g.,  revisiting a location it already visited or falsely assuming it is holding an object.
    \item \textbf{Short-sighted planning.} The ReAct agent typically exhibits short-sighted planning, making decisions that appear locally plausible but disregard the long-term task goal. 
\end{enumerate}
We recognize that {\em these two shortcomings stem from incorrect guidance for the agent’s reasoning process although the agent has better reasoning capability.} Indeed, for the same failure scenario in Figure~\ref{fig:when_thoughts_hinder_alfworld} left branch, we input a reasoning guidance sentence different from that of ReAct, 
to focus not on predicting the next step but  assess the agent’s current state in relation to the task goal, as seen in  Figure~\ref{fig:when_thoughts_hinder_alfworld} right branch. Now, it is seen that  the agent can better align its action with the overall objective by reflecting on what it currently knows and what it aims to achieve. This is also valid for failure cases in  
Figure~\ref{fig:when_thoughts_hinder_scienceworld} in Appendix \ref{appendix: another_thought_problem}.

\section{Proposed Method: ReflAct}

In the  LLM POMDP framework,  a thought $\tau_t \in \mathcal{T}$ should be generated so that it leads to the selection of an action $a_t$ that maximizes the expected long-term return when used to condition the action policy $\pi_\theta^{\text{act}}$, where the long-term return is defined as $G_t = \sum_{k=0}^\infty \gamma^k R_{t+k}$ with discount factor $\gamma \in [0, 1)$. Here, $R_{t+k}$ is the immediate reward at time $t+k$. Thus, the optimal thought $\tau_t^*$ can be defined as
$$
\tau_t^* = \mathop{\arg\max}_{\tau \in \mathcal{T}} \mathbb{E}_{a \sim \pi_\theta^{\text{act}}(\cdot \mid c_t \oplus \tau)} \left[ \mathbb{E}\left[ G_t \mid s_t, a \right] \right],
$$
where $c_t$ denotes the current context and $\oplus$ denotes concatenation.
To maximize the expected long-term return, the thought $\tau_t$ should not be generated from the observation solely to decide the next action. Instead, it should be formed with consideration of the final task goal at every time step, as illustrated in Figure~\ref{fig:direction_of_thought}. In this way, the agent can rely on thought for action generation without deviating from a goal-reaching path. Furthermore, since the environment is partially observable and the true state $s_t$ is not directly accessible, the agent must rely on an internal belief state inferred from its interaction history. So, the generated thought should make this belief more explicit and stable, while encoding task-relevant reasoning that effectively guides the downstream policy toward the long-term objective.

To achieve the generation of such thoughts, we do not use an additional module but exploit the LLM agent's reasoning capability maximally. For this,  we redesign a new thought generation instruction to better guide thought generation as follows:
\begin{tcolorbox}[
colback=white,
title=\small{Instruction for ReflAct},
colframe=gray,
coltitle=black,
colbacktitle=gray!10,
enhanced,
attach boxed title to top left = {yshift=-2mm}
]
\small
You should first \textbf{reflect on the agent's state in relation to the task goal}, and then output the action for this turn.
\end{tcolorbox}
\noindent In this new design, we have the explicit guiding terms `agent's state', `task goal' and their connection term `in relation to' to realize the desired state diagram of thought dependency in the lower part of  Figure  \ref{fig:direction_of_thought}.  
Following this instruction, the agent engages in reflective reasoning to gain a clearer understanding of both the current situation and the intended goal. We refer to so-generated thought as reflection and name the new reflection-action process 
 \textbf{ReflAct}.   
 To formalize this, we define a reflection space $K$, where each reflection $k \in K$ is a structured representation that aims to explicitly encode both the agent's internal belief state $M$ and the task goal $G$. The belief state captures the agent’s current understanding of the state based on its interaction history, whereas the goal component provides a concise summary of what the agent is trying to achieve. By explicitly representing both in the reflection, ReflAct allows the agent to condition its subsequent decision-making on a more coherent and goal-aligned internal context.
\begin{tcolorbox}[
  enhanced,
  title=\small{Example of Thought vs Reflection Before Action},
  colback=white,
  colframe=gray,
  coltitle=black, 
  colbacktitle=gray!10, 
  attach boxed title to top left = {yshift=-2mm},
  breakable
]
\small
\textcolor{thought}{Thought}: Now I find a spraybottle 2. Next, I need to take it.

\vspace{2mm}
\hrule
\vspace{2mm}

\textcolor{reflection}{Reflection}: Currently, I am at cabinet 2 and have found a spraybottle 2, which brings me closer to completing the task of placing it on the toilet.
\end{tcolorbox}
In addition, we incorporate one-shot examples based on the tasks from \citet{react, mpo} modified with Reflections on the agent’s state in relation to the task goal, generated by GPT-4o-mini \cite{gpt-4o}. Above, we present an example describing the modification from Thought to Reflection. The complete set of one-shot examples is provided in Appendix~\ref{appendix: prompts}.

\begin{table*}[t!]
\centering
\small
\renewcommand{\arraystretch}{1.5}
\caption{Performance comparison of ReflAct with NoThinking, ReAct, and Plan-and-Act across ALFWorld, ScienceWorld, and Jericho. SR and AR denote success ratio and average reward, respectively. Values in parentheses indicate percentage improvement over the NoThinking baseline.}
\begin{tabular}{p{2.3cm}p{2.2cm}C{1.5cm}C{1.05cm}C{1.05cm}C{1cm}C{1cm}C{2.2cm}}
\hline
\textbf{\multirow{2}{*}{Model}} & \textbf{\multirow{2}{*}{Prompting}} & \textbf{ALFWorld}  & \multicolumn{2}{c}{\textbf{ScienceWorld}} & \multicolumn{2}{c}{\textbf{Jericho}} & \textbf{\multirow{2}{*}{Average}} \\
\arrayrulecolor{gray!50}
\cline{3-7}
\arrayrulecolor{black} 
& & SR & AR & SR & AR & SR & \\
\hline
\hline
\multirow{4}{*}{GPT-4o}
& NoThinking 
& 76.1 
& 67.4 
& 50.2
& 27.8 
& 10.0 
& 46.3  \\
& ReAct 
& 85.1
& 68.7
& 55.9
& 50.4
& 20.0 
& 56.0 \small{(+21.0\%)}  \\
& Plan-and-Act
& 85.8
& 68.7
& 55.0
& 45.6
& 20.0
& 55.0 \small{(+18.8\%)} \\
& \cellcolor{reflact} \textbf{ReflAct} 
& \cellcolor{reflact} \textbf{93.3} 
& \cellcolor{reflact}\textbf{68.9} 
& \cellcolor{reflact} \textbf{57.8}
& \cellcolor{reflact} \textbf{53.2} 
& \cellcolor{reflact}\textbf{35.0} 
& \cellcolor{reflact}\textbf{61.6} \textcolor{blue}{\small{(+33.1\%)}}  \\
\hline

\multirow{4}{*}{GPT-4o-mini}
& NoThinking 
& 43.3 
& 42.3 
& 21.8
& 18.8 
& 5.0 
& 26.2 \\
& ReAct 
& 53.0
& 49.1
& \textbf{37.0}
& 29.8
& 15.0
& 36.8 \small{(+40.2\%)} \\
& Plan-and-Act
& 59.0
& 51.8
& 35.5
& 37.1
& 15.0
& 39.7 \small{(+51.1\%)} \\
& \cellcolor{reflact} \textbf{ReflAct} 
& \cellcolor{reflact} \textbf{66.4} 
& \cellcolor{reflact} \textbf{55.4} 
& \cellcolor{reflact} \textbf{37.0} 
& \cellcolor{reflact} \textbf{45.4} 
& \cellcolor{reflact} \textbf{20.0} 
& \cellcolor{reflact} \textbf{44.8} \textcolor{blue}{\small{(+70.9\%)}} \\
\hline

\multirow{4}{*}{\makecell[l]{Llama-3.1-8B\\-Instruct}}
& NoThinking 
& 21.6 
& 29.9 
& 14.2
& 10.0 
& 0.0 
& 15.1 \\
& ReAct 
& 29.1
& 43.0
& 27.5
& 12.8
& 0.0
& 22.5 \small{(+48.5\%)} \\
& Plan-and-Act
& 30.6
& 34.2
& 14.7
& 10.6
& 0.0
& 18.0 \small{(+19.0\%)} \\
& \cellcolor{reflact} \textbf{ReflAct} 
& \cellcolor{reflact} \textbf{60.5} 
& \cellcolor{reflact} \textbf{47.2} 
& \cellcolor{reflact} \textbf{33.2} 
& \cellcolor{reflact} \textbf{20.9} 
& \cellcolor{reflact} \textbf{10.0} 
& \cellcolor{reflact} \textbf{34.4} \textcolor{blue}{\small{(+126.9\%)}} \\
\hline

\multirow{4}{*}{\makecell[l]{Llama-3.1-70B\\-Instruct}}
& NoThinking
& 53.7 
& 65.6 
& 46.4
& 27.4 
& 5.0 
& 39.6 \\
& ReAct 
& 81.3
& 66.4
& 53.1
& 33.9 
& 10.0 
& 48.9 \small{(+23.5\%)} \\
& Plan-and-Act
& 81.3
& 67.7
& 52.6
& 29.6
& 10.0
& 48.2 \small{(+21.8\%)} \\
& \cellcolor{reflact} \textbf{ReflAct} 
& \cellcolor{reflact} \textbf{83.6} 
& \cellcolor{reflact} \textbf{73.7} 
& \cellcolor{reflact} \textbf{58.8}
& \cellcolor{reflact} \textbf{44.5} 
& \cellcolor{reflact} \textbf{20.0} 
& \cellcolor{reflact} \textbf{56.1} \textcolor{blue}{\small{(+41.6\%)}} \\
\hline
\end{tabular}
\label{table:main_results}
\vspace{-2mm}
\end{table*}

\begin{figure*}[h!]
    \centering
    \includegraphics[width=\linewidth]{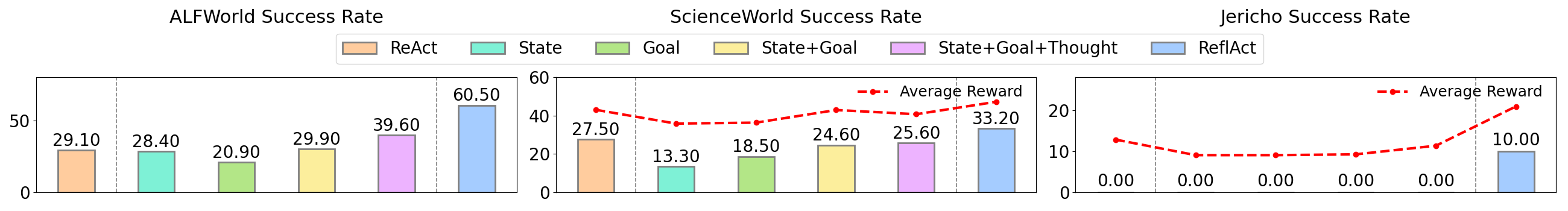}
    \caption{A comparison of ReAct, ReflAct, and various verbalization strategies (state, goal, state+goal, and state+goal with next-action reasoning), using Llama-3.1-8B-Instruct as the agent model. Bars represent success rate; the red dashed line indicates average reward.}
    \label{fig: ablation}
    \vspace{-4mm}
\end{figure*}

\section{Related Works}

Building on CoT reasoning~\cite{cot}, numerous approaches  attempted to enhance LLM agent capability. ReAct~\cite{react} pioneered the interleaving of reasoning with action steps for interactive environments, while  Plan-and-Solve~\cite{pos} introduced higher-level planning in the initial stage. Recent enhancement modules  sought to improve ReAct's CoT through various mechanisms: Reflexion~\cite{reflexion} through post-task analysis of failures \textit{after task completion}, WKM~\cite{wkm}, MPO~\cite{mpo}, and DC~\cite{dc} via external memory of environment knowledge, and RAFA~\cite{rafa} by generating and evaluating future possible trajectory predictions. However, when the ReAct backbone incorrectly grounds past observations, all these approaches suffer from degraded performance. Therefore, ReflAct retouches the reasoning process itself to enable more grounded and goal-aligned decision making. In our experiments, we will validate the importance of this approach compared to the design of enhancement modules, as well as its synergy with them. Additional related works are discussed in the Appendix~\ref{expanded_rel_work}.

\section{Experiments}

\subsection{Experimental Settings}
\noindent \textbf{Benchmarks.} ~~ We benchmarked ReflAct on three widely used text-based environments: ALFWorld \cite{alfworld}, ScienceWorld \cite{scienceworld}, and Jericho \cite{jericho}. ALFWorld evaluates embodied agents on household tasks, ScienceWorld assesses procedural and scientific reasoning in educational scenarios, and Jericho tests agents in classic interactive fiction games that require commonsense reasoning and memory. For evaluation, ALFWorld uses binary task success, while ScienceWorld and Jericho provide dense reward signals, enabling evaluation based on both success rate and average reward, calculated as the mean reward across all tasks. Additional benchmark details are provided in Appendix \ref{appendix: benchmark}.

\noindent \textbf{Agent Models.} ~ We use GPT-4o and GPT-4o-mini \cite{gpt-4o}, and Llama-3.1-8B/70B-Instruct \cite{llama3} as underlying models. GPT-4o variants serve as proprietary models, while Llama-3.1-Instruct variants represent open-source counterparts, each with large and small sizes.

\subsection{Comparison with Prior Methods} \label{sec: main results}

\noindent \textbf{Baselines.} ~~ We first compare ReflAct with three reasoning frameworks. Details for each can be found in Appendix~\ref{appendix: backbone baseline}.

\noindent (1) NoThinking \cite{NoThinking}: The agent generates an action directly at each time step  without any reasoning step.

\noindent (2) ReAct \cite{react}: The agent first reasons about the next action at each time step and then generates an action.

\noindent (3) Plan-and-Act: In addition to NoThinking and ReAct, we designed another reasoning framework inspired by Plan-and-Solve \cite{pos} for comparison. Here, the agent is instructed at the first timestep ($t=1$) to `plan your approach to the task, and then output the action.' It performs reasoning only at this step and  outputs actions without further thoughts in subsequent timesteps.

\noindent \textbf{Results.} ~~ Table \ref{table:main_results} presents the performance of ReflAct compared with the NoThinking, ReAct, and Plan-and-Act baselines across ALFWorld, ScienceWorld, and Jericho. As shown in the table, the ReAct agent outperforms the NoThinking agent, highlighting the effectiveness of the \textit{Thinking before Action} framework. Plan-and-Act, which begins with a single planning step and then executes actions directly, performs comparably or slightly better than ReAct in stronger models like GPT-4o. This suggests that when a model has enough capacity, initial goal-based reasoning is effective, and flawed intermediate reasoning may offer little benefit. However, in weaker models like Llama, Plan-and-Act underperforms because these models struggle to retain and follow the initial plan. In contrast, ReflAct, which reflects on both the state and the goal at every timestep, consistently outperforms all the baseline reasoning frameworks regardless of backbone model or task. Notably, ReflAct achieved performance improvements of 77.9\%, 25.9\%, and 101.1\% over NoThinking in ALFWorld, ScienceWorld, and Jericho, respectively, and \textbf{improvements of 36.4\%, 8.5\%, and 38.1\% over ReAct in the same environments.}

\noindent \textbf{Importance of Reflection over Simply Stating.} ~~ To verify the source of ReflAct’s superior performance gain, we conducted an ablation study. ReflAct reflects on the agent’s state in relation to the task goal, so we compared this with variants in which the agent simply verbalizes its state, its goal, or both, without engaging in reflection. The instructions used in the ablation studies can be found in Appendix \ref{apendix: ablation setting}.

As shown in Figure \ref{fig: ablation}, we evaluated four settings: the agent verbalizes (1)  the current state, (2) the task goal, (3) both the state and the goal, and (4) the state and goal while also reasoning about the next action. The results indicate that merely stating the state or goal underperforms compared to the ReAct baseline. Adding an explicit next-action thought in addition to verbalizing the state and goal yields performance gains in ALFWorld compared to ReAct, but still falls short of ReAct in ScienceWorld and Jericho. In contrast, ReflAct, which explicitly reflects on the relationship between the state and the goal, demonstrates a substantially greater performance gain. These findings suggest that reflection, specifically reasoning about the state in the context of the goal, is more effective than simply verbalizing the state or the goal.

\subsection{ReflAct with Enhancement Modules} \label{sec: enhancement_modules}

We discuss ReflAct alongside existing enhancement modules. We consider Reflexion \cite{reflexion}, the use of post-task reflection in  ReAct to improve planning, and WKM \cite{wkm}, which incorporates a parametric world model into ReAct. Discussion on  RAFA \cite{rafa}, which focuses on long-sighted planning via next-state prediction, and MPO \cite{mpo}, which improves planning through online preference learning, can be found in Appendix \ref{appendix: discussion-enhancement}.

\subsubsection{ReflAct and Reflexion}

Reflexion~\cite{reflexion} performs post-task reflection by analyzing failed trajectories and generating improvement plans for future trials. Since post-task reflection operates independently of the underlying reasoning process, we analyze its impact when combined with three reasoning frameworks: NoThinking, ReAct, and ReflAct.

\begin{figure}[h!]
  \centering
  \includegraphics[width=0.48\textwidth]{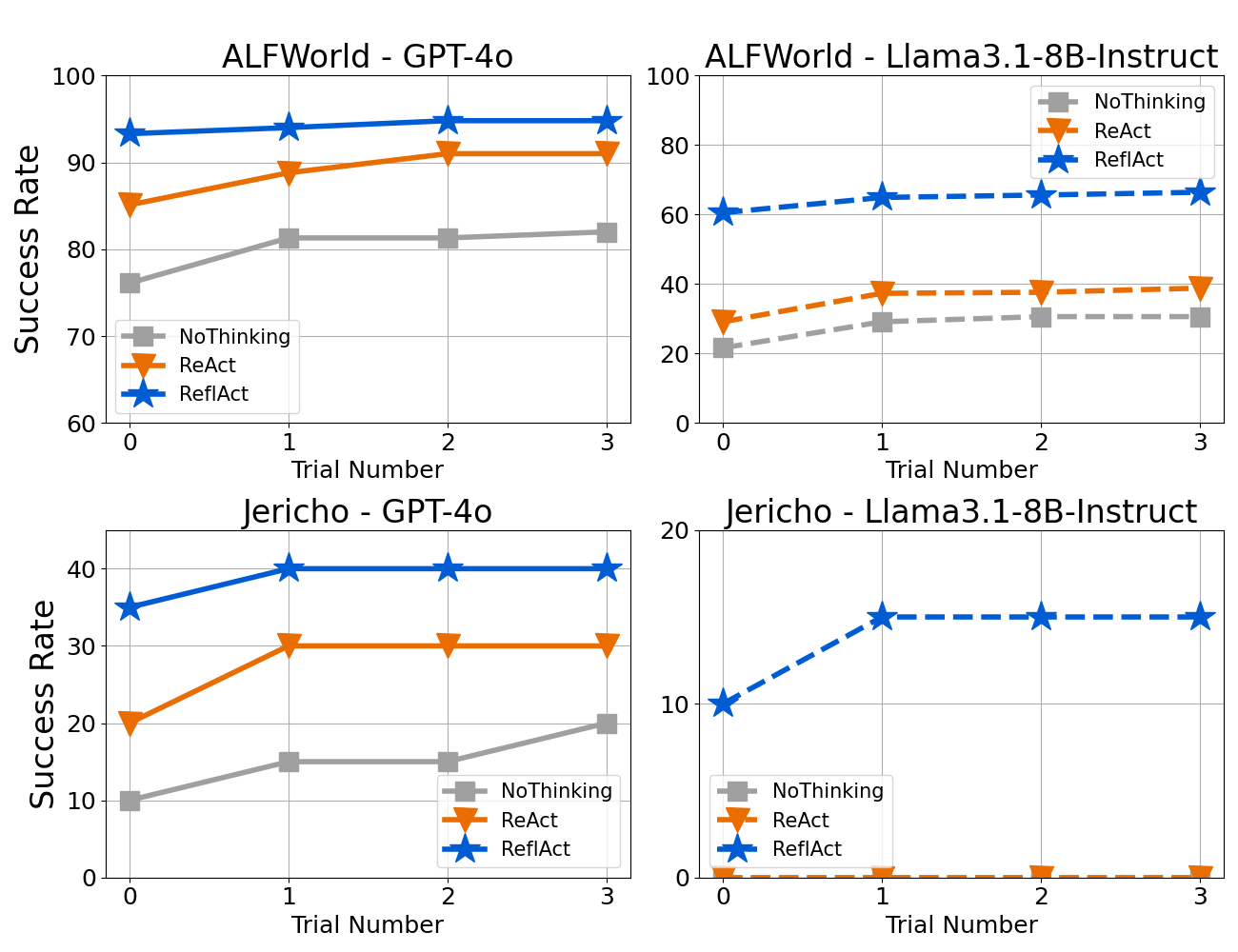}
  \caption{Success rate change with Reflexion post-task reflection (NoThinking, ReAct, ReflAct). }
  \label{fig:reflexion}
  \vspace{-4mm}
\end{figure}

Figure \ref{fig:reflexion} shows the results of applying Reflexion to each backbone in  ALFWorld and Jericho. We followed \citet{reflexion} in applying post-task reflection after each trial, repeating this for three trials. The results show that post-task reflection generally improves performance across all base agents, except Llama NoThinking and ReAct in Jericho. However, even after applying Reflexion, the NoThinking and ReAct agents still fall short of ReflAct's trial 0 performance. This highlights that when the initial reasoning backbone is weak, even repeated feedback across trials provides only limited benefit. Although post-task reflection aids reasoning by analyzing failures, enhancing in-task reflection through a stronger backbone proves more fundamental. Notably, the GPT-4o ReflAct agent with Reflexion achieved an unprecedented 94.8\% success rate in ALFWorld, bringing household robots one step closer to reality, given sufficient hardware support.

\subsubsection{ReflAct and WKM} \label{sec: external_problem}

WKM \cite{wkm} improves LLM agents' decision-making in interactive environments by using Task Knowledge for global planning and State Knowledge for local context. Its parametric knowledge model, fine-tuned on knowledge-augmented trajectories, enables context-aware decision-making by referencing past similar situations. WKM learns and utilizes an \textit{external} knowledge model in contrast to ReflAct, which intends to perform \textit{internal} state modeling within the reasoning backbone.

We applied WKM to ReAct and further examined a scenario where state knowledge is placed immediately before  action, replacing the thought, to  guide the action. This allows us to assess the impact of explicit state knowledge compared to ReflAct’s internal state modeling.
\begin{figure}[h!]
  \centering
  \includegraphics[width=0.48\textwidth]{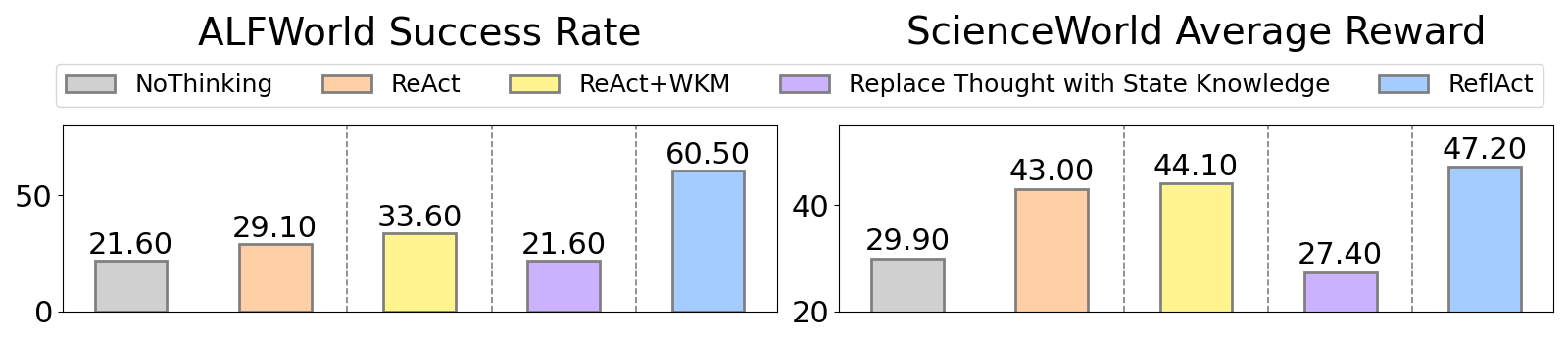}
  \vspace{-0.5cm}
  \caption{Performance: ReflAct, NoThinking, ReAct, and ReAct+WKM with thought replaced by WKM state knowledge (Llama-3.1-8B-Instruct as agent)}
  \label{fig:wkm}
  \vspace{-3mm}
\end{figure}
Figure \ref{fig:wkm} shows performance related to WKM. Applying WKM to ReAct results in a performance improvement, but it still falls short of ReflAct. Replacing the thought step with state knowledge to directly guide the action actually degrades performance, highlighting the side effect of incorrect guidance from the external model. This suggests that encouraging the model to perform internal state modeling on its own can be more effective than relying on state knowledge generated by an external model.

\subsection{Further Discussion on ReflAct's  Ability} \label{sec: reflact_effects}

In this section, we analyze how ReflAct successfully solves tasks, and in Appendix \ref{appendix: discussion-reflact}, we compare the hallucinated action ratio and token length between ReflAct and baseline reasoning frameworks.

\subsubsection{Failure Ratio Comparison Between Reasoning Frameworks}

Building on the discussion of Figure \ref{fig:fail_plot} in Section \ref{sec: ungrounded_thought_problem}, we analyze the proportion of tasks failed by the NoThinking, ReAct, and ReflAct agents across 134 tasks in the ALFWorld domain.

\begin{figure}[h!]
  \centering
  \vspace{-0.3cm}
  \includegraphics[width=0.237\textwidth]{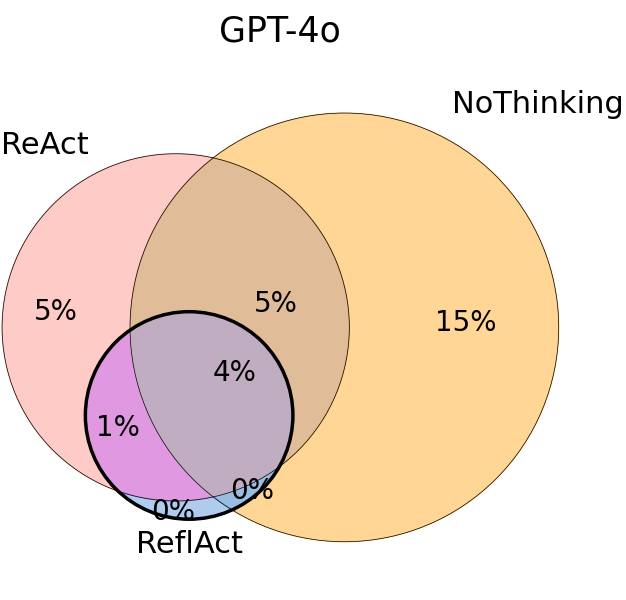}
  \hfill
  \includegraphics[width=0.237\textwidth]{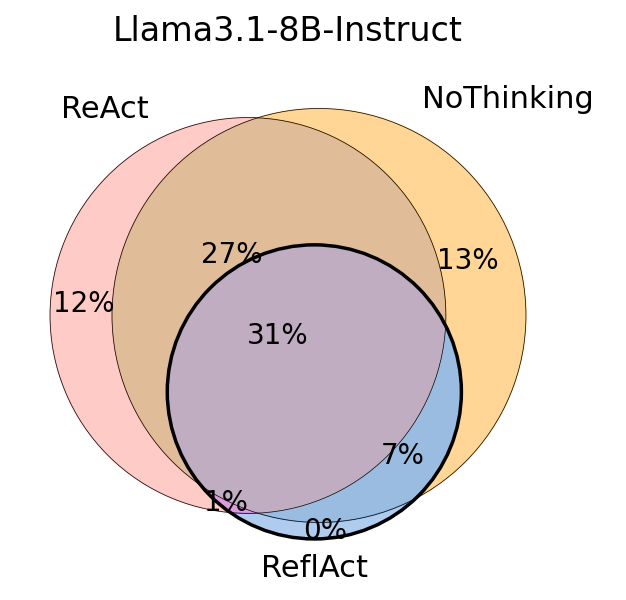}
  \caption{Failure ratio comparison diagram of NoThinking, ReAct, and ReflAct on 134 ALFWorld tasks, using GPT-4o and Llama-3.1-8B-Instruct as agent models.}
  \vspace{-0.3cm}
  \label{fig:fail_plot_3}
\end{figure}


Figure \ref{fig:fail_plot_3} illustrates the failure distribution among the three agents. Consistent with earlier observations, the NoThinking agent accounts for the largest number of unique failures, especially in tasks where it alone fails. The ReAct agent, while incorporating reasoning, introduces additional failure cases not observed in NoThinking. This suggests that ungrounded or incorrect reasoning can lead to novel and potentially harmful errors. Notably, there are no tasks in which only the ReflAct agent fails. ReflAct fails only in cases where other agents also fail, and does not introduce new failure modes. This highlights the benefit of reflection grounded in the agent's state and task goal, suggesting that reflective reasoning contributes to safer and more reliable decision-making in LLM agents.

\subsubsection{ReflAct Self-Corrects Automatically} \label{sec: case2}

LLM agents, especially when operating in unfamiliar environments with only partially observable information, are  prone to making mistakes. Therefore, it is essential that they possess the ability to self-correct by identifying the cause of their errors and adjusting their behavior accordingly to return to a correct course of action. The example below by a ReflAct agent demonstrates that ReflAct maintains an internal belief over the current state and continuously aligns its decision-making with the task goal, enabling the agent to {\em recognize mistakes and flexibly recover from them}, ultimately guiding itself back onto the correct path.

\sethlcolor{reflact} 

\vspace{-0.1cm}
\begin{tcolorbox}[
  colback=white,
  colframe=gray,
  enhanced,
  fontupper=\scriptsize\fontfamily{pcr}\selectfont
]
\footnotesize{\textbf{Your task is cool some lettuce and put it in countertop.}} \\\\
\scriptsize{... \\
\textcolor{action}{Action}:   take lettuce 1 from countertop 2 \\
\textcolor{obs}{Obs}:   You pick up the lettuce 1 from the countertop 2. \\ \\
\textcolor{reflection}{Reflection}:   I am at countertop 2, holding lettuce 1. I need to cool the lettuce before putting it on the countertop to complete the task. \\
\textcolor{action}{Action}:   cool lettuce 1 with fridge 1 \\
\textcolor{obs}{Obs}:   Nothing happens. \\ \\
\textcolor{reflection}{Reflection}:    I am at countertop 2, holding lettuce 1. \hl{\textbf{The action to cool the lettuce with the fridge did not work because I need to be at the fridge to perform that action.}}  \\
\textcolor{action}{Action}:   go to fridge 1 \\
\textcolor{obs}{Obs}:   The fridge 1 is open. In it, ... \\ \\
\textcolor{reflection}{Reflection}:    I am at fridge 1, holding lettuce 1. \hl{\textbf{I can cool the lettuce now that I'm at the fridge.}}   \\
\textcolor{action}{Action}: cool lettuce 1 with fridge 1 \\
\textcolor{obs}{Obs}: You cool the lettuce 1 using the fridge 1. \\
...}
\end{tcolorbox}

\section{Conclusion}

We proposed ReflAct, a reflection-based reasoning framework that addresses key limitations of ReAct. While ReAct focuses on next-action thinking, it often fails to maintain coherent internal beliefs about the environment and goal, leading to poor decisions in complex, partially observable tasks. In contrast, ReflAct emphasizes ongoing reflection on state and goal, enabling more grounded and reliable decision making. Our experiments show that ReflAct consistently outperforms ReAct and other baselines, demonstrating the value of reflection in long-horizon reasoning. We hope this work inspires further research into more robust and informative reasoning guidance for intelligent agents.

\section*{Limitations}

While ReflAct shows strong performance in interactive agent environments, its applicability to other domains such as mathematics and coding remains unexplored. These tasks pose unique challenges: mathematics requires step-by-step deduction, while programming demands consistency in logic, state, and syntax. Extending ReflAct to these areas may require adapting its reflection mechanism to better align the current state with task-specific goals. Incorporating structured reasoning that explicitly supports goal-state alignment could enhance performance, and we consider this a promising direction for future work.

In addition, as discussed in Appendix \ref{appendix: discussion-reflact}, ReflAct results in a modest increase in token length compared to ReAct. To address this, it is worth exploring more efficient formulations that retain the benefits of state-goal reflection while reducing verbosity. Possible approaches include compressing reflection outputs, using more concise representations of reasoning steps, or integrating summarization modules to maintain informativeness within a smaller token budget.

Finally, there is growing interest in enhancing reasoning with external modules. Incorporating such components into ReflAct could further improve its reflection process. For example, an external verifier might evaluate reflected goals or suggest fixes for inconsistencies. Exploring such hybrid architectures remains a promising direction for future research.

\section*{Ethical Statements}
While LLM Agents represent powerful tools with significant practical applications, they also pose substantial risks if misused or deployed without proper safeguards. Of particular concern is their tendency to hallucinate or generate ungrounded reasoning, which can lead to unreliable or potentially harmful decisions in real-world contexts. This research aims to address these challenges by developing more grounded decision-making frameworks that enhance reliability and reduce hallucination. By focusing on improving the alignment between the agent's internal beliefs and the actual environment state, our work contributes to the broader goal of creating more trustworthy AI systems that can safely operate in complex, partially observable environments while maintaining consistent goal-directed behavior.

\section*{Acknowledgements}
This work was supported in part by Institute of Information \& Communications Technology Planning \& Evaluation (IITP) grant funded by the Korea government (MSIT) (No. RS-2024-00457882, AI Research Hub Project) and in part by the National Research Foundation of Korea(NRF) grant funded by the Korea government(MSIT) (No. RS-2025-02263628).

\bibliography{custom}

\appendix

\normalsize
\section{Expanded Related Works}
\label{expanded_rel_work}

\textbf{Belief-State Modeling in RL.} ~~  Reinforcement Learning (RL) has aimed to endow agents with an internal state representation that captures all relevant information from past observations~\cite{rodriguez1999reinforcement}. In partially observable environments, this often takes the form of a belief state, a probability distribution over possible world states, which is updated as new observations are received. \citet{kaelbling1998planning} describe planning as tracking a sufficient statistic of the history (i.e., the belief), which serves as the agent’s core state for action selection. This approach has been further developed in deep RL, where agents learn to track belief states using recurrent neural networks~\cite{hausknecht2015deep}, memory-augmented architectures~\cite{oh2016control}, and world models~\cite{ha2018world}. Various studies~\cite{dreamer, muzero, hafner2025mastering} have demonstrated the effectiveness of belief state modeling, enabling more effective planning and reliable decision-making.

\noindent \textbf{State Representations in LLMs.} ~~ Recent work on LLMs has investigated how internal or external state representations can support reasoning and planning, analogous to belief-state modeling in RL. A prominent line of research, inspired by Code-as-Policies~\cite{codeaspolicy}, takes an extrinsic approach: structuring observations into symbolic formats (e.g., textual summaries, PDDL, finite state machines) and injecting them into prompts~\cite{hao2023rap, shirai2024vilain, wu2024stateflow}. While this enhances grounding and interpretability in robotics, it often depends on predefined code or external modules. ReflAct extends this direction by prompting LLMs to generate and reason over goal-conditioned state representations directly, without auxiliary modules, enabling more generalized and efficient decision-making.

\noindent  \textbf{Backbone Reasoning Frameworks.}  ~~ The emergence of LLMs has introduced new agent design paradigms, especially via prompting-based reasoning strategies. CoT prompting~\cite{cot} showed that models can handle complex tasks by outlining intermediate steps. ReAct~\cite{react} extended this to interactive settings by interleaving reasoning and action, while Plan-and-Solve~\cite{pos} added high-level planning before stepwise execution. However, recent studies~\cite{NoThinking} argue that these CoT-based approaches are often token-inefficient and prone to hallucinations. We attribute these reliability issues to the absence of an explicit belief state. ReflAct addresses this by maintaining and reflecting belief states to guide reasoning and improve both efficiency and reliability.

\noindent \textbf{Enhancement Reasoning Modules.} ~~ Recently, a growing body of research has explored add-on or post-processing modules that operate on top of reasoning backbones to enhance reasoning capabilities. Reflexion~\cite{reflexion} improves performance by analyzing failed trajectories and storing insights for future use. WKM~\cite{wkm} mitigates hallucinations by grounding decisions in a learned world model that reflects realistic dynamics. RAFA~\cite{rafa} strengthens the reasoning-acting link by imagining future trajectories and evaluating them with separate \textit{Model} and \textit{Critic} components, forming a structured loop of planning, action, and feedback. Unlike these approaches, ReflAct is not layered on top of a reasoning backbone but instead redefines the backbone itself, enabling more efficient reasoning.

\noindent \textbf{Imitation Learning-Based Approaches.} ~~ In addition, many studies enhance agents' reasoning abilities through imitation learning. For instance, ETO~\cite{eto} performs imitation learning on a 7B model using expert trajectories and then applies direct preference optimization (DPO)~\cite{dpo} with both successful and failed trajectories. SwiftSage~\cite{swiftsage} applies imitation learning to a smaller 770M model but still relies on GPT-4 or GPT-3.5-Turbo for action extraction in switching scenarios, thereby maintaining dependence on large models. More recently, KnowSelf~\cite{knowself} enhances LLMs’ knowledge utilization by constructing self-awareness training data: categorizing reasoning into Fast, Slow, and Knowledgeable Thinking, regenerating reasoning with prompts tailored to each type, and then fine-tuning an LLM on it. In contrast, ReflAct requires neither gold trajectories nor parameter updates, and can be directly applied to both large and small, closed- or open-source models, enabling robust generalization to unseen tasks.

\section{Another Case of the Ungrounded Thought Problem} \label{appendix: another_thought_problem} 

\begin{figure*}[t!]
\centering
  \includegraphics[width=\textwidth]{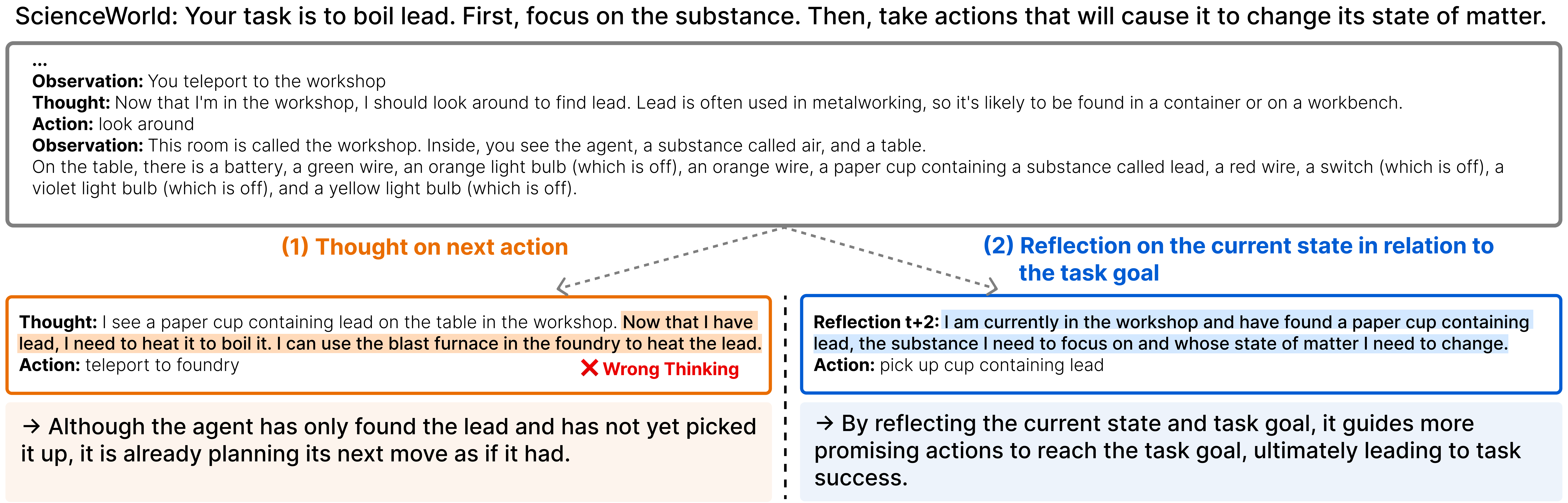}
  \caption{Another failure case of ReAct, following Figure \ref{fig:when_thoughts_hinder_alfworld}, is seen in the ScienceWorld task boil lead. The agent finds the lead in the workshop. (1) It mistakenly believes it is holding the lead, proceeds with the task, and fails. (2) After reflecting on its current state in relation to the task goal, the agent correctly recognizes its situation, takes goal-aware actions, and successfully completes the task.}
  \label{fig:when_thoughts_hinder_scienceworld}
\end{figure*}

In addition to the discussion of the ReAct agent's failure case in Figure~\ref{fig:when_thoughts_hinder_alfworld} in Section~\ref{sec: ungrounded_thought_problem}, we examine another failure case in ScienceWorld~\cite{scienceworld}. As shown in Figure~\ref{fig:when_thoughts_hinder_scienceworld} (1), the agent is tasked with boiling lead. To do this, the agent must find lead, transfer it to a suitable heat source, and then boil it. During the previous steps, the agent successfully located the lead in the workshop. Now, the agent must carry it to the foundry. However, the agent mistakenly assumes that it is already holding the lead and proceeds with the next action based on this false assumption, ultimately resulting in a hallucination.


In the same situation, similar to the discussion in Section~\ref{sec: ungrounded_thought_problem}, we replace the thought with a reflection sentence about the current state and task goal, as seen in the right branch of Figure~\ref{fig:when_thoughts_hinder_scienceworld}. This allows the agent to clearly recognize the current situation, make goal-aware long horizon decisions, and ultimately succeed in the task.

\section{Benchmark Details} \label{appendix: benchmark}

\begin{itemize}[itemsep=0pt, leftmargin=*]
    \item \textbf{ALFWorld} \cite{alfworld}: 
    ALFWorld is a text-based benchmark derived from the ALFRED dataset, designed to evaluate an agent's ability to interpret and execute natural language instructions in interactive, multi-step household tasks. It converts embodied vision-and-language tasks into textual form, allowing for assessment of high-level planning, language grounding, and sequential decision making in a simulated environment.

    \item \textbf{ScienceWorld} \cite{scienceworld}: 
    ScienceWorld is a text-based benchmark for evaluating agents on scientific reasoning and procedural task completion. Grounded in standardized K–12 science curricula, it tests an agent’s ability to follow instructions, perform virtual experiments, make causal inferences, and manipulate objects across domains such as physics, chemistry, and biology. Tasks require multi-step interaction, hypothesis testing, and language grounding in a dynamic environment.

    \item \textbf{Jericho} \cite{jericho}: 
    Jericho is a suite of text-based adventure games, each with distinct environments and roles. Tasks require completing a sequence of implicit subgoals, often in specific order. Since these subgoals often do not follow intuitive causal logic, agents must rely on interaction to model the environment and accurately infer their current state.
    
\end{itemize}

In ALFWorld, success is measured using a binary signal: agents receive a reward of 1 only when the task is fully completed, and 0 otherwise. In contrast, the official environments of ScienceWorld and Jericho provide both a success flag and a progress reward, which assigns a value between 0 and 1 based on the degree of task completion. However, we discovered a bug in both ScienceWorld and Jericho where the success flag could be incorrectly set even when the progress reward was low or incomplete. For example, in Jericho, a task was sometimes marked as successful even when the agent died during execution. To address this, we adjusted the success criteria: a task was considered successful only if the progress reward was at least 0.7 in ScienceWorld, or exactly 1.0 in Jericho.

In our experiments, the setups for ALFWorld and ScienceWorld were based on the official implementation by MPO \cite{mpo} (\url{https://github.com/WeiminXiong/MPO}), and we used the same test set for evaluation. The ALFWorld test set consists of 134 tasks across six different household task types, including `pick-and-place', `pick-heat-then-place', and `pick-two-and-place'. The ScienceWorld test set includes 211 tasks across 24 different scientific experiment types, including `Boil', `Test Conductivity of Unknown Substances', and `Grow Fruit'.

The experiments for Jericho were based on the implementation by AgentBoard \cite{agentboard} (\url{https://github.com/hkust-nlp/AgentBoard}). The Jericho test set consists of 20 tasks, each corresponding to a different game, including 905, Acorncourt, Afflicted, Balances, Dragon, Jewel, Library, Omniquest, Reverb, Snacktime, Zenon, Zork1, Zork2, Zork3, Detective, Night, Pentari, Weapon, Darkhunt, and Loose.

\section{Details of Experiment Setup}


\subsection{Experimental Setup: Comparison of Backbone Reasoning Frameworks} \label{appendix: backbone baseline}

We use the same environment code base and test set for all reasoning frameworks, including ReflAct. For specific environments, we follow existing implementations: ALFWorld and ScienceWorld are based on the implementation by MPO \cite{mpo}, and Jericho is based on the implementation by AgentBoard \cite{agentboard}, as described in detail in the section above. We implement baseline reasoning frameworks using the following instructions:

\begin{itemize}[itemsep=0pt, leftmargin=*]
\item Instruction for NoThinking
\begin{tcolorbox}[
  colback=white,
  colframe=gray,
  enhanced
]
\small
You should \textbf{directly output the action} in this turn. 
\end{tcolorbox}

\item Instruction for ReAct

\begin{tcolorbox}[
  colback=white,
  colframe=gray,
  enhanced
]
\small
You should first think about the current condition and \textbf{plan for your future actions, and then output your action} in this turn. 
\end{tcolorbox}

Note that this ReAct instruction is from \citet{eto, mpo}.

\item Instruction for Plan-and-Act
\begin{tcolorbox}[
  colback=white,
  colframe=gray,
  enhanced
]
\small
You should first think about the given task and \textbf{plan your approach to the task, and then output the action} for this turn. 
\end{tcolorbox}

The one-shot example for ReAct is based on \citet{react, mpo} and can be found in Appendix~\ref{appendix: prompts}. For NoThinking, we remove the thought step from all timesteps in the same example. For Plan-and-Act, we retain only the thought from the first timestep ($t=1$). 
\end{itemize}

\subsection{Experimental Setup: Comparison of Enhancement Reasoning Modules}

In our experiments, we discuss four enhanced reasoning modules: Reflexion \cite{reflexion} and WKM \cite{wkm} in Section~\ref{sec: enhancement_modules}, and RAFA \cite{rafa}, MPO \cite{mpo}, and ReflAct in Appendix~\ref{appendix: discussion-enhancement}. We implemented each module based on the official repositories provided by the original papers. For Reflexion, we used the official implementation on top of each environment's base code: ALFWorld and ScienceWorld from \citet{mpo}, and Jericho from \citet{agentboard}. In addition, for WKM, we used the publicly available dataset on Hugging Face (\url{https://huggingface.co/collections/zjunlp/wkm-6684c611102213b6d8104f84}) to reproduce the original setup and conducted experiments using WKM's world model.
\begin{itemize}[itemsep=0pt, leftmargin=*]
    \item \textbf{Reflexion} \cite{reflexion}: \url{https://github.com/noahshinn/reflexion}
    \item \textbf{WKM} \cite{wkm}: \url{https://github.com/zjunlp/WKM}
    \item \textbf{RAFA} \cite{rafa}: \url{https://github.com/agentification/RAFA_code}
    \item \textbf{MPO} \cite{mpo}: \url{https://github.com/WeiminXiong/MPO}
\end{itemize}


\subsection{Experimental Setup: ReflAct Ablation Study} \label{apendix: ablation setting}

In the ReflAct ablation study (Figure~\ref{fig: ablation}, Section~\ref{sec: main results}), we used the following instructions.


\begin{itemize}[itemsep=0pt, leftmargin=*]
\item Instruction for State
\begin{tcolorbox}[
  colback=white,
  colframe=gray,
  enhanced
]
\small
You should first \textbf{consider the current state, then output the action} for this turn.\\
Your output must strictly follow this format: \\"State: current state.\textbackslash n Action: your next action".
\end{tcolorbox}


\item Instruction for Goal
\begin{tcolorbox}[
  colback=white,
  colframe=gray,
  enhanced
]
\small
You should first \textbf{consider the task goal, then output the action} for this turn. \\
Your output must strictly follow this format: \\"Goal: task goal.\textbackslash n Action: your next action".
\end{tcolorbox}


\item Instruction for State+Goal
\begin{tcolorbox}[
  colback=white,
  colframe=gray,
  enhanced
]
\small
You should first \textbf{consider the current state and task goal, then output the action} for this turn.\\
Your output must strictly follow this format: \\"State: current state, Goal: task goal \textbackslash n Action: your next action".
\end{tcolorbox}


\item Instruction for State+Goal+Thought
\begin{tcolorbox}[
  colback=white,
  colframe=gray,
  enhanced
]
\small
You should first \textbf{consider the task goal, current state, and plan for your future actions, then output the action} for this turn.\\
Your output must strictly follow this format: \\"Goal: task goal.\textbackslash n Current location: your current location.\textbackslash n Current Inventory: your current inventory.\textbackslash n Thought: your thoughts.\textbackslash n Action: your next action".
\end{tcolorbox}

State+Goal+Thought extends State+Goal by adding ReAct-style reasoning. An example of State+Goal+Thought before action is:
\begin{tcolorbox}[
  colback=white,
  colframe=gray,
  enhanced
]
\small

Goal: put some spraybottle on toilet
\\
Current location: cabinet 2
\\
Current inventory: none
\\
Thought: Now I find a spraybottle 2. Next, I need to take it.
\\
Action: take spraybottle 2 from cabinet 2
\end{tcolorbox}
\end{itemize}

\section{Resources}
For inference with the Llama-3.1-Instruct model, we used one A6000 48GB GPU for the 8B model and two A100 80GB GPUs for the 70B model. For training the WKM world model, we used four A6000 48GB GPUs.

\section{More Discussion on Enhancement Modules} \label{appendix: discussion-enhancement}

\subsection{ReflAct and RAFA}

RAFA \cite{rafa} introduces a next-state prediction framework where a language model recursively expands a tree of possible future action-observation trajectories, selecting actions based on the highest estimated value. While this foresight aids long-term reasoning, ReflAct offers practical advantages in efficiency and grounded decision-making by internally modeling the current state.

\begin{figure}[h!]
    \centering
    \includegraphics[width=\linewidth]{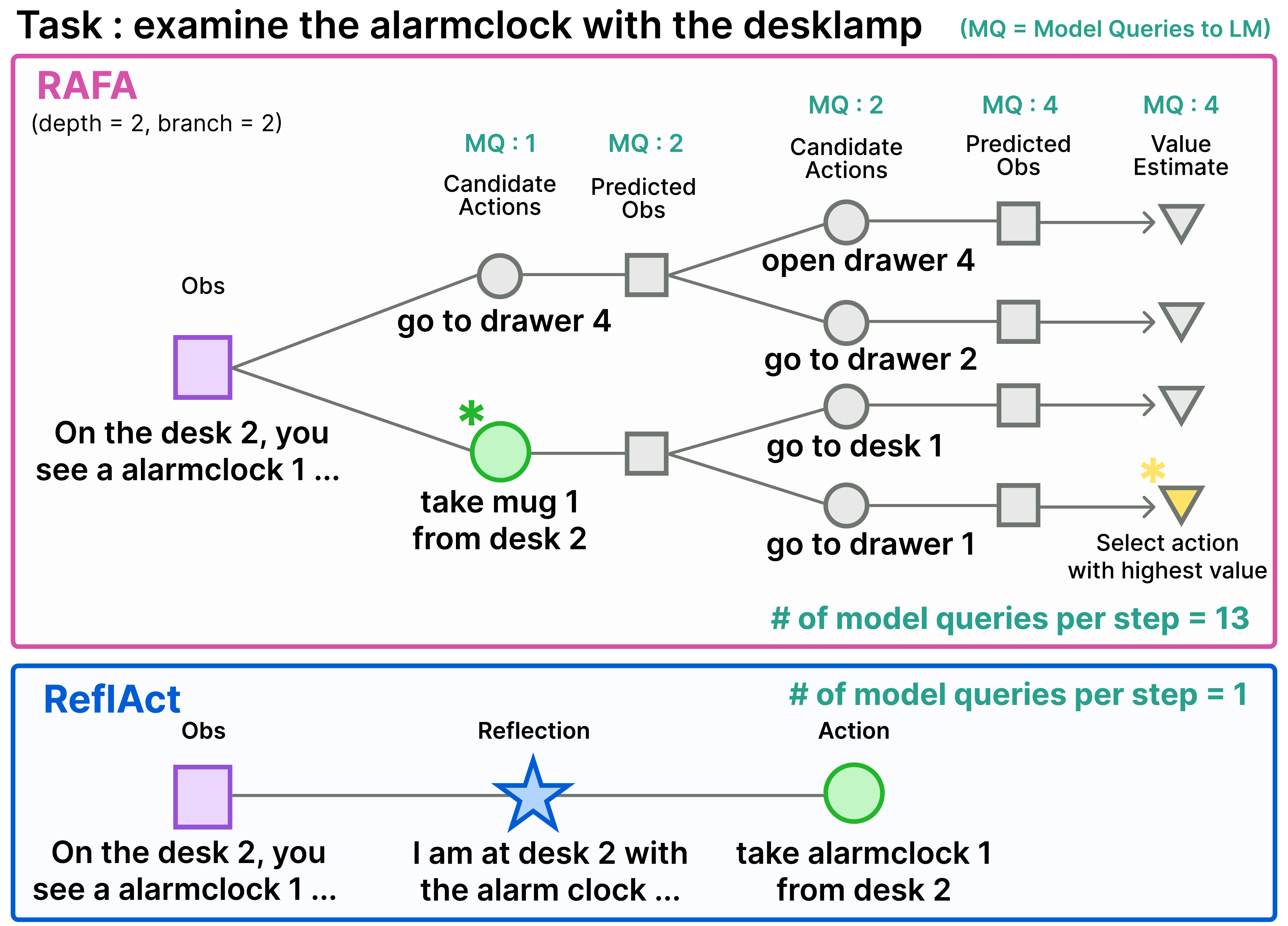}
    \caption{Comparison between RAFA and ReflAct.}
    \label{fig: rafa}
\end{figure}

As shown in Figure~\ref{fig: rafa}, RAFA constructs a tree at each time step $t$, requiring 13 queries to the language model per step with depth $d=2$ and branching factor $b=2$: 3 for action sampling, 6 for observation prediction, and 4 for value estimation. This overhead grows rapidly with depth due to exponential node expansion. In contrast, ReflAct needs only a single query per step to jointly produce a reflection and action, offering constant cost and greater efficiency when latency or throughput is constrained.

Figure~\ref{fig: rafa} also highlights a behavioral difference between the two methods. Given the task \textit{"examine the alarmclock with the desklamp"} in ALFWorld, both agents observe an alarmclock on desk 2. ReflAct grounds its reflection in this state and selects the correct action, while RAFA relies on predicted trajectories that miss the action and lead to an unrelated exploratory move. This shows that ReflAct aligns decisions more closely with the actual state, achieving more reliable behavior with significantly lower cost via a single LLM call per step.

\subsection{ReflAct and MPO}

Recently, \citet{mpo} proposed a framework that enhances LLM agents by generating and refining high-level meta plans based on agent feedback. The meta plans are initially learned through supervised fine-tuning on expert-generated examples and subsequently optimized via preference-based learning \cite{dpo} using feedback from agent executions. The resulting meta planner can be applied to various agents in a plug-and-play manner without requiring additional retraining. Accordingly, we applied the publicly available ALFWorld MPO meta plans from \url{https://github.com/WeiminXiong/MPO} to NoThinking, ReAct, and ReflAct, and compared the resulting performance gains.

\begin{figure}[h!]
    \centering
    \includegraphics[width=\linewidth]{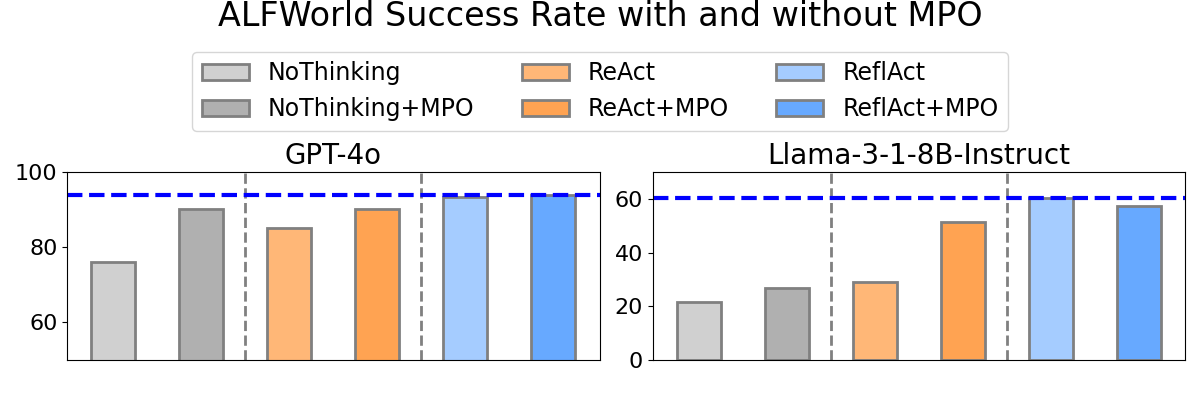}
    \caption{Performance comparison across NoThinking, ReAct, and ReflAct when applying MPO meta-plan.}
    \label{fig: mpo}
    \vspace{-4mm}
\end{figure}

As shown in Figure \ref{fig: mpo}, applying MPO meta plans to NoThinking and ReAct leads to substantial performance improvements. However, notably, even with these gains, NoThinking+MPO and ReAct+MPO still fall short of ReflAct’s performance. This further reinforces previous findings that strengthening the reasoning backbone itself is more crucial than simply adding enhancement modules to ReAct.

Applying MPO meta plans to ReflAct results in slight performance gains for GPT-4o, but a minor performance drop for Llama-3.1-8B-Instruct. This may be due to the fact that the meta plans were trained on tasks different from those in the test set; in some out-of-distribution cases, they may offer misleading guidance that negatively impacts the agent’s behavior. The limited effectiveness of meta plans on ReflAct also suggests that its grounded decision-making already enables high-quality planning.

\section{Evaluating ReflAct with Advanced Reasoning Models} \label{appendix:qwen3}
\begin{table}[h!]
\centering
\small
\renewcommand{\arraystretch}{1.3}
\begin{tabular}{p{2cm}C{1.35cm}C{1.3cm}C{1.35cm}}
\hline
\textbf{Metric} & \textbf{NoThinking} & \textbf{ReAct} & \textbf{ReflAct} \\
\hline
\hline
ALFWorld SR        & 60.5 & 65.7 & \cellcolor{reflact}\textbf{80.6} \\
\hline
ScienceWorld AR    & 45.2 & 50.3 & \cellcolor{reflact}\textbf{51.8} \\
ScienceWorld SR    & 26.5 & 34.1 & \cellcolor{reflact}\textbf{36.0} \\
\hline
Jericho AR         & 32.1 & 30.4 & \cellcolor{reflact}\textbf{33.2} \\
Jericho SR         & 10 & 5 & \cellcolor{reflact}\textbf{20} \\
\hline
\end{tabular}
\caption{Performance of Qwen3-8B with ReflAct with NoThinking, and ReAct across ALFWorld, ScienceWorld, and Jericho. SR and AR denote success ratio and average reward, respectively.}
\label{table:qwen_results}
\vspace{-2mm}
\end{table}

Recently released models such as Qwen3 (May 2025) \cite{qwen3} have a model size comparable to Llama-3.1-8B-Instruct used in the main results (Table ~\ref{table:main_results}), while offering stronger reasoning capability. We investigate whether ReflAct maintains its effectiveness on such models. 

Since the Qwen3 series supports inference-time scaling (also referred to as “thinking mode”), we employed the Qwen3-8B model, which has stronger reasoning ability. However, in multi-turn LLM agentic benchmarks, the use of excessively long reasoning traces often prevented the model from properly producing both the \textit{Thinking} and \textit{Action} outputs. Therefore, we disabled the inference-time scaling mode and compared the performance of NoThinking, ReAct, and ReflAct across ALFWorld, ScienceWorld, and Jericho.

When comparing the results of Table~\ref{table:main_results} with those of Qwen3-8B (Table~\ref{table:qwen_results}), we find that Qwen3-8B demonstrates overall stronger performance across all three reasoning strategies, achieving results comparable to or even surpassing those of Llama-3.1-8B-Instruct and, in some cases, GPT-4o-mini. Within the Qwen3 results, ReflAct shows marginal improvements over ReAct in ScienceWorld, but exhibits significant success rate gains in ALFWorld and Jericho. This indicates that even for LLMs with strong single-turn reasoning ability, applying them to long-horizon LLM agent tasks still requires aligning the agent’s belief state with both the actual environment state and the final goal, and further highlights the effectiveness of ReflAct.

\section{Additional Benchmark Evaluation} \label{appendix:balrog}

BabyAI~\cite{babyai} is a natural language-based 2D grid-world where an LLM agent navigates rooms, perceives objects through stepwise observations, and generates text-based actions to complete tasks. It includes five navigation tasks (“open,” “pickup,” “put next,” “pick up seq go to,” and “goto”) that test the agent’s ability to explore reliably and demonstrate spatial reasoning over long horizons.

We conducted experiments on all five BabyAI tasks, with 10 episodes each (50 in total), limited to a maximum of 128 steps per episode, and used average progression as the evaluation metric. Using the official Balrog implementation~\cite{balrog} (\url{https://github.com/balrog-ai/BALROG}), we evaluated ReAct and ReflAct on four models (GPT-4o, GPT-4o-mini, Llama-3.1-8B-Instruct, and Qwen3-8B without thinking mode), with results presented in Table~\ref{table:balrog}.

\begin{table}[h!]
\centering
\small
\renewcommand{\arraystretch}{1.3}
\begin{tabular}{p{3.3cm}C{1.5cm}C{1.5cm}}
\hline
\textbf{Model} & \textbf{ReAct} & \textbf{ReflAct} \\
\hline
\hline
GPT-4o            & 64.0 & \cellcolor{reflact}\textbf{76.0} \\
GPT-4o mini       & 40.0 & \cellcolor{reflact}\textbf{48.0} \\
Llama-3.1-8B-Instruct & 38.0 & \cellcolor{reflact}\textbf{50.0} \\
Qwen3-8B          & 32.0 & \cellcolor{reflact}\textbf{36.0} \\
\hline
\end{tabular}
\caption{Performance of ReAct and ReflAct on the BabyAI benchmark across GPT-4o, GPT-4o-mini, Llama-3.1-8B-Instruct, and Qwen3-8B.}
\label{table:balrog}
\vspace{-2mm}
\end{table}

In all models we used, ReflAct outperforms ReAct. These results demonstrate that ReflAct consistently improves performance across models of different scales, highlighting its robustness as a reasoning framework.

\section{More Discussion on ReflAct's Effects} \label{appendix: discussion-reflact}

\subsection{ReflAct Prevents Suboptimality Through Goal-Oriented Reasoning} \label{sec: case1}

\begin{figure}[h!]
  \centering
  \includegraphics[width=0.48\textwidth]{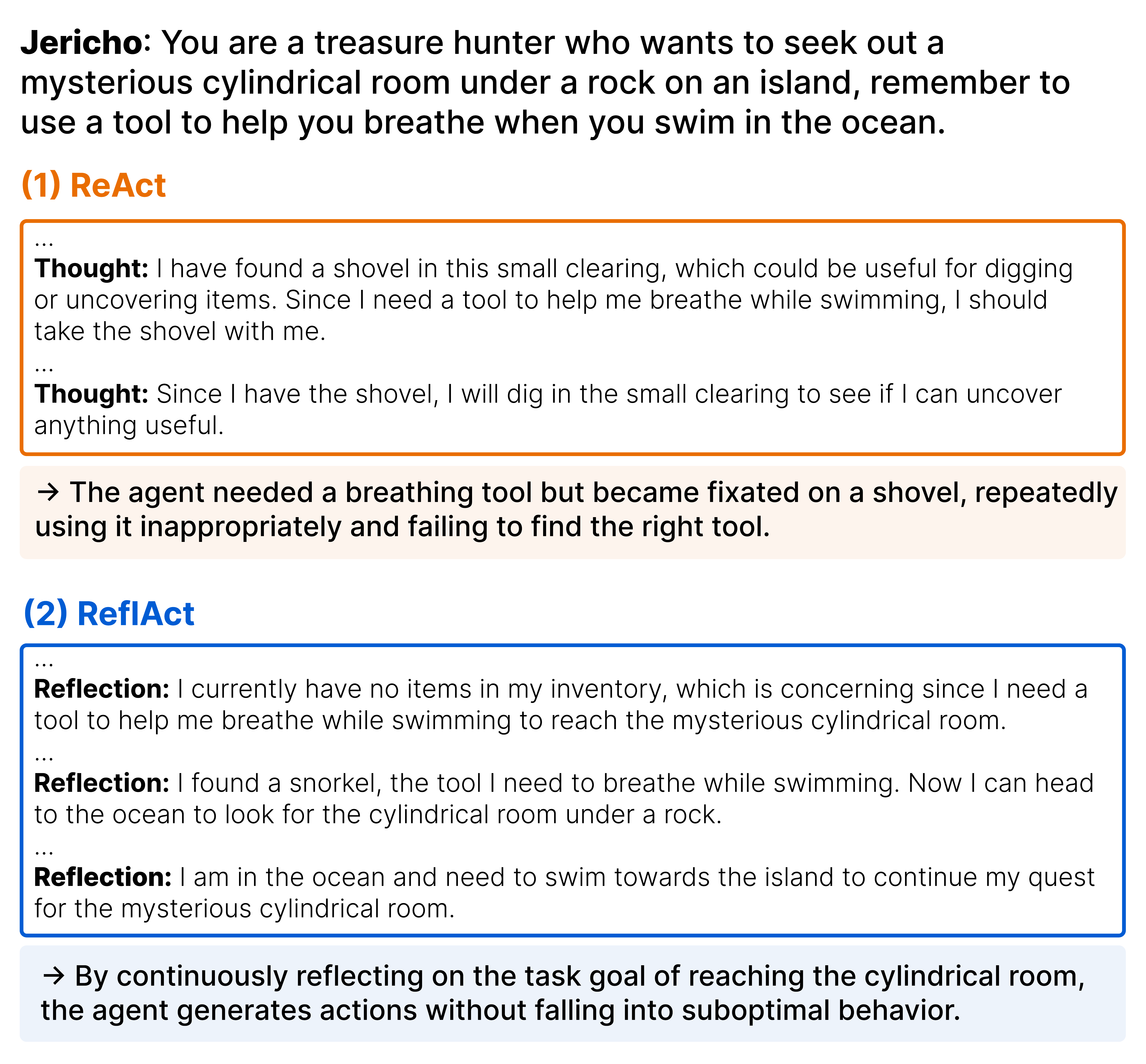}
  \caption{Comparison of ReAct's Thought and ReflAct's Reflection on a Jericho task: ReAct fixates on a found shovel, losing sight of the main goal and failing; ReflAct succeeds through goal-oriented reasoning.}
  \label{fig:case1_jericho}
\end{figure}

ReflAct enables long-sighted planning by continuously reflecting on the current state and goal. Figure~\ref{fig:case1_jericho} illustrates this with an example in a task where the agent must gather swimming-related items and ultimately reach a cylindrical room. As seen, ReAct fails due to short-sighted reasoning, whereas ReflAct successfully achieves the goal through consistent reflection on the state and objective.

\subsection{Robustness of ReflAct across Prompt Variants}

We also conducted experiments to examine whether ReflAct, as a prompt-based method, remains robust when using different prompt variants. Specifically, we tested two types of variations: (1) instruction paraphrasing and (2) ICL example paraphrasing.

Below, we provide the original instruction along with three revised instructions tested with GPT-4o-mini (for the full original prompt, see Appendix~\ref{appendix: prompts}). The results in Table~\ref{table:revised_inst} show that among the three revisions, Instruction 2—which introduced the most substantial changes—led to the largest performance drop, but the effect remained minor fluctuations without any significant impact overall.

\begin{itemize}[itemsep=0pt, leftmargin=*]
\item Original instruction
\begin{tcolorbox}[
  colback=white,
  colframe=gray,
  enhanced
]
\small
You should first reflect in one sentence on the agent's state in relation to the task goal, and then output the action for this turn.
\end{tcolorbox}

\item Revised instruction 1
\begin{tcolorbox}[
  colback=white,
  colframe=gray,
  enhanced
]
\small
Reflect in a one sentence on what the agent’s state indicates about achieving the goal, and then output the action for this turn.
\end{tcolorbox}

\item Revised instruction 2
\begin{tcolorbox}[
  colback=white,
  colframe=gray,
  enhanced
]
\small
First, write a short sentence that explains the agent’s progress or position relative to the objective, and then output the action for this turn.
\end{tcolorbox}

\item Revised instruction 3
\begin{tcolorbox}[
  colback=white,
  colframe=gray,
  enhanced
]
\small
Offer a concise sentence that reflects on the agent’s current state in the context of the goal, and then output the action for this turn.
\end{tcolorbox}

\end{itemize}

\begin{table}[h!]
\centering
\small
\renewcommand{\arraystretch}{1.3}
\begin{tabular}{p{4.5cm}C{2.3cm}}
\hline
\textbf{Variants} & \textbf{ALFWorld SR} \\
\hline
\hline
ReflAct (original)   & 66.4 \\
ReflAct (revised 1)  & 65.7 \\
ReflAct (revised 2)  & 64.1 \\
ReflAct (revised 3)  & 65.7 \\
\hline
\end{tabular}
\caption{Performance comparison of GPT-4o-mini under instruction paraphrasing in ALFWorld ReflAct.}
\label{table:revised_inst}
\vspace{-2mm}
\end{table}

In addition, we performed experiments by paraphrasing the ICL examples provided in Appendix~\ref{appendix: prompts}, where we revised two segments of the reasoning part, as shown below. The experiments were conducted with GPT-4o, GPT-4o-mini, Llama-3.1-8B-Instruct, and Llama-3.1-70B-Instruct.

\begin{itemize}[itemsep=0pt, leftmargin=*]
\item Original icl example
\begin{tcolorbox}[
  colback=white,
  colframe=gray,
  enhanced
]
\small
Currently, I am at cabinet 2, not holding anything, but the closed cabinet hinders me from finding the spraybottle needed to place on the toilet.\\
...\\
Currently I am located at cabinet 2, now holding a spraybottle 2, and I am ready to complete the task of placing the spraybottle on the toilet, as I have obtained spraybottle 2.
\end{tcolorbox}

\item Revised icl example
\begin{tcolorbox}[
  colback=white,
  colframe=gray,
  enhanced
]
\small
Currently located at cabinet 2, the agent is not holding anything, and the situation is challenging as the cabinet is closed, hindering the search for a spray bottle necessary to complete the task of placing it on the toilet.\\
...\\
Currently located at cabinet 2, the agent is now holding a spray bottle, and the situation has become favorable for completing the task as the agent has successfully found the spray bottle needed to place it on the toilet.
\end{tcolorbox}



\end{itemize}

\begin{table}[h!]
\centering
\small
\renewcommand{\arraystretch}{1.3}
\begin{tabular}{p{3.1cm}C{1.6cm}C{1.6cm}}
\hline
\textbf{Model} & \textbf{ReflAct (original)} & \textbf{ReflAct (revised)} \\
\hline
\hline
GPT-4o               & 93.3 & 91.0 \\
GPT-4o-mini          & 66.4 & 65.7 \\
Llama-3.1-8B-Instruct  & 60.5 & 62.7 \\
Llama-3.1-70B-Instruct & 83.6 & 83.6 \\
\hline
\end{tabular}
\caption{Comparison of ReflAct performance between the original and revised icl examples.}
\label{table:revised_icl}
\vspace{-2mm}
\end{table}

As shown in Table~\ref{table:revised_icl}, while the other models experienced a performance drop, Llama-3.1-8B-Instruct exhibited a slight improvement. This suggests that ReflAct serves as the primary determinant of performance as a reasoning framework, while the optimal prompt may vary depending on the model. Furthermore, consistent with the results in Table~\ref{table:revised_inst}, the fact that the variants exhibit only very small performance differences demonstrates that ReflAct remains robust to prompt paraphrasing.

\subsection{Token Length vs. Hallucination Action Rate in ReAct and ReflAct}

Here, we further compare ReflAct and ReAct in terms of token length and the hallucinated action rate. Specifically, token length refers to the number of tokens in each thought (or reflection) with action generated by the model, normalized by the number of steps to account for variations in agent trajectory length. The hallucinated action rate is defined as the proportion of actions that result in erroneous observations (e.g., "Nothing happens"). Both metrics were averaged across the four agent models used in our experiments: GPT-4o, GPT-4o-mini, Llama-3.1-8B/70B-Instruct.

\begin{figure}[h!]
    \centering
    \includegraphics[width=\linewidth]{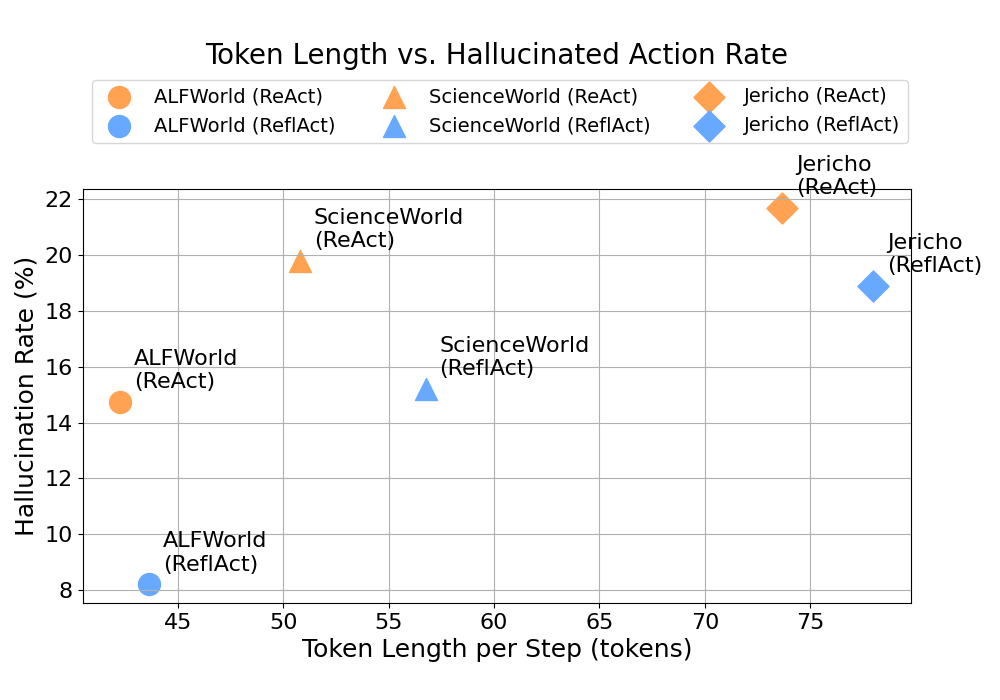}
    \caption{Comparison of the average token length and average hallucination action rate between ReAct and ReflAct in ALFWorld, ScienceWorld, and Jericho.}
    \label{fig: token_length}
\end{figure}

Figure \ref{fig: token_length} shows that ReflAct produces slightly longer token sequences than ReAct, as it incorporates reflection on the current state and goal into its reasoning process. Interestingly, despite the increased token length, the hallucination rate decreases. This suggests that hallucinations in LLM agents may not arise simply from increased reasoning (i.e., overthinking), but rather from unstructured or unguided reasoning processes.

\subsection{Number of Reasoning Steps Comparison}

In Table~\ref{table:number_of_steps}, we summarize the average number of reasoning steps required for multi-turn task solving across benchmarks from the main results (Table~\ref{table:main_results}), averaged over the four agent models used in our experiments: GPT-4o, GPT-4o-mini, and Llama-3.1-8B/70B-Instruct.

\begin{table}[h!]
\centering
\small
\renewcommand{\arraystretch}{1.3}
\begin{tabular}{p{2cm}C{1.4cm}C{1.3cm}C{1.35cm}}
\hline
\textbf{Environment} & \textbf{NoThinking} & \textbf{ReAct} & \textbf{ReflAct} \\
\hline
\hline
ALFWorld      & 21.0 & 18.6 & \cellcolor{reflact}\textbf{16.5} \\
ScienceWorld    & 24.0 & 20.3 & \cellcolor{reflact}\textbf{19.8} \\
Jericho       & 28.4 & 26.2 & \cellcolor{reflact}\textbf{24.6} \\
\hline
\end{tabular}
\caption{Number of reasoning steps used in NoThinking, ReAct, and ReflAct across ALFWorld, ScienceWorld, and Jericho.}
\label{table:number_of_steps}
\vspace{-2mm}
\end{table}

As shown in Table~\ref{table:number_of_steps}, the average reasoning steps across all domains (ALFWorld, ScienceWorld, and Jericho) follow the order NoThinking > ReAct > ReflAct, with ReflAct requiring the fewest steps. This demonstrates that ReflAct guides the LLM agent to task success much more quickly and once again underscores the importance of goal-state reflection in ReflAct for embodied tasks.

\section{ReflAct Success \& Failure Case Study}

To more deeply analyze when ReflAct succeeds or fails, we additionally examined the success rates by the six tasks in ALFWorld (Put, Clean, Heat, Cool, Examine, Put Two).

\begin{table}[h!]
\centering
\small
\renewcommand{\arraystretch}{1.3}
\begin{tabular}{p{1.6cm}C{2.5cm}C{2.5cm}}
\hline
\textbf{Task} & \textbf{GPT-4o} & \textbf{Llama-3.1-8B-Instruct} \\
\hline
\hline
Put      & 95.8 & 83.3 \\
Clean    & 96.8 & \textbf{38.7} \\
Heat     & 78.3 & 56.5 \\
Cool     & 95.2 & 66.7 \\
Examine  & 100.0 & 72.2 \\
Put Two  & 94.1 & 52.9 \\
\hline
Average  & 93.3 & 60.5 \\
\hline
\end{tabular}
\caption{Success rate of ReflAct by ALFWorld task.}
\label{table:alfworld_task}
\vspace{-2mm}
\end{table}

Looking at the results, we see that ReflAct particularly struggled with the Heat task on GPT-4o and the Clean task on Llama-3.1-8B-Instruct. We therefore additionally analyzed the cases where ReflAct failed on the GPT-4o Heat task and the Llama-3.1-8B-Instruct ALFWorld Clean task. The Heat and Clean tasks require the agent to pick up the target object, perform the appropriate action at the designated location (heating at the microwave or cleaning at the sink basin), and then place the object at the goal location to complete the task.

There are two main reasons why GPT-4o ReflAct fails on the Heat task:
\begin{itemize}[itemsep=0pt, leftmargin=*]
    \item The agent fails to find the target object within the allotted time → an exploration problem.
    \item In the task “put a hot cup in cabinet,” the agent needs to find and pick up a cup, but instead it first finds a mug and tries to use it to complete the task, ultimately failing. This shows the agent does not understand the environment rule that it must specifically find a cup.
\end{itemize}

The main reason Llama-3.1-8B-Instruct ReflAct fails on the Clean task is that it misunderstands the environment’s rule. In this task, cleaning only requires performing the clean action at the sink basin—no materials like soap are needed. Since the agent is explicitly instructed that it can use the clean action, it should simply execute that action at the sink basin to succeed.

However, as shown in the example below, smaller models like Llama-3.1-8B sometimes attempt to clean using other tools instead of trying the clean action first, which leads to failure.

\begin{tcolorbox}[
colback=white,
colframe=gray,
coltitle=black,
colbacktitle=gray!10,
enhanced,
]
\small
Reflection: Currently, I am at sinkbasin 1 and holding a soapbottle, while the tasks of cleaning and putting a soapbar in the cabinet are still pending because I did not find a soapbar at the sinkbasin. I should try another location to find a soapbar. I can check one by one, starting with cabinet 2.
\end{tcolorbox}

ReflAct is designed to serve as an enhanced backbone reasoning module that demonstrates improved decision-making capabilities on its own, without requiring additional layered reasoning modules. We expect that incorporating a memory-based enhancement module tailored for ReflAct, which can further learn the environment’s rules through trial and error, could make it an even more powerful reasoning system. 

\newpage
\onecolumn
\section{Prompt} \label{appendix: prompts}

\subsection{ALFWorld}
\subsubsection{Instructions}

\vspace{0.5cm}
\noindent
\centering
\begin{tcolorbox}[
  colback=white,
  colframe=gray,
  enhanced,
  fontupper=\small\fontfamily{pcr}\selectfont,
  sharp corners=southwest,
  coltitle=black,
  colbacktitle=gray!10,
  title = {\small{Instruction for ALFWorld Nothinking, ReAct, ReflAct}}
]
Interact with a household to solve a task. Imagine you are an intelligent agent in a household environment and your target is to perform actions to complete the task goal. At the beginning of your interactions, you will be given the detailed description of the current environment and your goal to accomplish. \\
For each of your turn, you will be given the observation of the last turn. You should directly output the action in this turn. Your output must strictly follow this format:"Action: your next action". \\
\textcolor{nothinking}{(Nothinking)} For each of your turn, you will be given the observation of the last turn. You should directly output the action in this turn. Your output must strictly follow this format:"Action: your next action". \\
\textcolor{thought}{(ReAct)} For each of your turn, you will be given the observation of the last turn. You should first think about the current condition and plan for your future actions, and then output your action in this turn. Your output must strictly follow this format:"Thought: your thoughts.\textbackslash n Action: your next action". \\
\textcolor{reflection}{(ReflAct)} For each of your turn, you will be given the observation of the last turn. You should first reflect in one sentence on the agent's state in relation to the task goal, and then output the action for this turn. Your output must strictly follow this format:"Reflection: your reflection.\textbackslash n Action: your next action".\\
The available actions are:\\
1. go to {recep}\\
2. take {obj} from {recep}\\
3. put {obj} in/on {recep}\\
4. open {recep}\\
5. close {recep}\\
6. use {obj}\\
7. clean {obj} with {recep}\\
8. heat {obj} with {recep}\\
9. cool {obj} with {recep}\\
where {obj} and {recep} correspond to objects and receptacles.\\
After your each turn, the environment will give you immediate feedback based on which you plan your next few steps. if the envrionment output "Nothing happened", that means the previous action is invalid and you should try more options.\\
Reminder: \\
The action must be chosen from the given available actions. Any actions except provided available actions will be regarded as illegal.
\end{tcolorbox}
\captionof{figure}{Instruction for ALFWorld Nothinking, ReAct, ReflAct}

\newpage

\subsubsection{ICL Examples}
\vspace{0.5cm}
\begin{flushleft}
One of our baselines, NoThinking, is a mode that does not involve the reasoning process of ReAct or ReflAct. Accordingly, the ICL examples for NoThinking omit the \textit{thought} and \textit{reflection} components of ReAct and ReflAct, and consist only of a chain of observations and actions.
\vspace{0.5cm}

\end{flushleft}
\noindent
\centering
\begin{tcolorbox}[
  colback=white,
  colframe=gray,
  enhanced,
  fontupper=\small\fontfamily{pcr}\selectfont,
  sharp corners=southwest,
  coltitle=black,
  colbacktitle=gray!10,
  title = {\small{Prompt for ALFWorld ReAct, "pick and place" task}},
]

You are in the middle of a room. Looking quickly around you, you see a cabinet 4, a cabinet 3, a cabinet 2, a cabinet 1, a countertop 1, a garbagecan 1, a handtowelholder 2, a handtowelholder 1, a sinkbasin 2, a sinkbasin 1, a toilet 1, a toiletpaperhanger 1, and a towelholder 1.\\
Your task is to: put some spraybottle on toilet. \\\\
\textcolor{thought}{Thought}: To solve the task, I need to find and take a sparybottle, then put it on toilet. First I need to find a spraybottle. A spraybottle is more likely to appear in cabinet 1-4, countertop 1, toilet 1, sinkbasin 1-2, garbagecan 1. I can check one by one, starting by going to cabinet 1.\\
\textcolor{action}{Action}: go to cabinet 1 \\
\textcolor{obs}{Observation}: On the cabinet 1, you see a cloth 1, a soapbar 1, a soapbottle 1. \\
\textcolor{action}{Action}: go to cabinet 2 \\
\textcolor{obs}{Observation}: The cabinet 2 is closed. \\
\textcolor{action}{Action}: open cabinet 2 \\
\textcolor{obs}{Observation}: You open the cabinet 2. The cabinet 2 is open. In it, you see a candle 1, and a spraybottle 2. \\
\textcolor{thought}{Thought}: Now I find a spraybottle 2. Next, I need to take it.\\
\textcolor{action}{Action}: take spraybottle 2 from cabinet 2 \\
\textcolor{obs}{Observation}: You pick up the spraybottle 2 from the cabinet 2. \\
\textcolor{thought}{Thought}: Now I take a spraybottle 2. Next, I need to put it in/on toilet 1.\\
\textcolor{action}{Action}: go to toilet 1 \\
\textcolor{obs}{Observation}: On the toilet 1, you see a soapbottle 2. \\
\textcolor{action}{Action}: put spraybottle 2 in/on toilet 1 \\
\textcolor{obs}{Observation}: You put the spraybottle 2 in/on the toilet 1.
\end{tcolorbox}
\captionof{figure}{Prompt for ALFWorld ReAct, "pick and place" task}

\vspace{1cm}

\noindent
\centering
\begin{tcolorbox}[
  colback=white,
  colframe=gray,
  enhanced,
  fontupper=\small\fontfamily{pcr}\selectfont,
  sharp corners=southwest,
  coltitle=black,
  colbacktitle=gray!10,
  title = {\small{Prompt for ALFWorld ReflAct, "pick and place" task}},
  breakable
]

You are in the middle of a room. Looking quickly around you, you see a cabinet 4, a cabinet 3, a cabinet 2, a cabinet 1, a countertop 1, a garbagecan 1, a handtowelholder 2, a handtowelholder 1, a sinkbasin 2, a sinkbasin 1, a toilet 1, a toiletpaperhanger 1, and a towelholder 1.\\
Your task is to: put some spraybottle on toilet. \\\\
\textcolor{reflection}{Reflection}: To solve the task, I need to find and take a sparybottle, then put it on toilet. First I need to find a spraybottle. A spraybottle is more likely to appear in cabinet 1-4, countertop 1, toilet 1, sinkbasin 1-2, garbagecan 1. I can check one by one, starting by going to cabinet 1.\\
\textcolor{action}{Action}: go to cabinet 1 \\
\textcolor{obs}{Observation}: On the cabinet 1, you see a cloth 1, a soapbar 1, a soapbottle 1. \\
\textcolor{reflection}{Reflection}: Currently, I am at cabinet 1, not holding anything, and am searching for a spraybottle to place on the toilet, but instead I find a cloth, a soapbar, and a soapbottle.\\
\textcolor{action}{Action}: go to cabinet 2 \\
\textcolor{obs}{Observation}: The cabinet 2 is closed. \\
\textcolor{reflection}{Reflection}: Currently, I am at cabinet 2, not holding anything, but the closed cabinet hinders me from finding the spraybottle needed to place on the toilet.\\
\textcolor{action}{Action}: open cabinet 2 \\
\textcolor{obs}{Observation}: You open the cabinet 2. The cabinet 2 is open. In it, you see a candle 1, and a spraybottle 2. \\
\textcolor{reflection}{Reflection}: Currently, I am at cabinet 2 and have found a spraybottle 2, which brings me closer to completing the task of placing it on the toilet.\\
\textcolor{action}{Action}: take spraybottle 2 from cabinet 2 \\
\textcolor{obs}{Observation}: You pick up the spraybottle 2 from the cabinet 2. \\
\textcolor{reflection}{Reflection}: Currently I am located at cabinet 2, now holding a spraybottle 2, and I am ready to complete the task of placing the spraybottle on the toilet, as I have obtained spraybottle 2.\\
\textcolor{action}{Action}: go to toilet 1 \\
\textcolor{obs}{Observation}: On the toilet 1, you see a soapbottle 2. \\
\textcolor{reflection}{Reflection}: Currently, I am at toilet 1, still holding spraybottle 2, and the task is nearly complete, with only the placement remaining.\\
\textcolor{action}{Action}: put spraybottle 2 in/on toilet 1 \\
\textcolor{obs}{Observation}: You put the spraybottle 2 in/on the toilet 1.
\end{tcolorbox}
\captionof{figure}{Prompt for ALFWorld ReflAct, "pick and place" task}

\vspace{1cm}

\noindent
\centering
\begin{tcolorbox}[
  colback=white,
  colframe=gray,
  enhanced,
  fontupper=\small\fontfamily{pcr}\selectfont,
  sharp corners=southwest,
  coltitle=black,
  colbacktitle=gray!10,
  title = {\small{Prompt for ALFWorld ReAct, "pick clean then place" task}},
]

You are in the middle of a room. Looking quickly around you, you see a cabinet 13, a cabinet 12, a cabinet 11, a cabinet 10, a cabinet 9, a cabinet 8, a cabinet 7, a cabinet 6, a cabinet 5, a cabinet 4, a cabinet 3, a cabinet 2, a cabinet 1, a coffeemachine 1, a countertop 1, a diningtable 1, a drawer 1, a fridge 1, a garbagecan 1, a microwave 1, a shelf 3, a shelf 2, a shelf 1, a sinkbasin 1, a stoveburner 4, a stoveburner 3, a stoveburner 2, a stoveburner 1, and a toaster 1.\\
Your task is to: put a clean lettuce in diningtable. \\\\
\textcolor{thought}{Thought}: To solve the task, I need to find and take a lettuce, then clean it with sinkbasin, then put it in diningtable. First I need to find a lettuce. A lettuce is more likely to appear in fridge 1, diningtable 1, sinkbasin 1, stoveburner 1-3, cabinet 1-13. I can check one by one, starting by going to fridge 1.\\
\textcolor{action}{Action}: go to fridge 1 \\
\textcolor{obs}{Observation}: The fridge 1 is closed.\\
\textcolor{action}{Action}: open fridge 1\\
\textcolor{obs}{Observation}: You open the fridge 1. The fridge 1 is open. In it, you see a cup 3, a egg 2, a potato 3, and a potato 2.\\
\textcolor{action}{Action}: go to diningtable 1\\
\textcolor{obs}{Observation}: On the diningtable 1, you see a apple 1, a bread 1, a butterknife 2, a cup 2, a fork 2, a knife 2, a knife 1, a ladle 1, a lettuce 1, a mug 2, a mug 1, a pan 2, a peppershaker 1, a spatula 3, a tomato 2, and a tomato 1. \\
\textcolor{thought}{Thought}: Now I find a lettuce 1. Next, I need to take it.\\
\textcolor{action}{Action}: take lettuce 1 from diningtable 1 \\
\textcolor{obs}{Observation}: You pick up the lettuce 1 from the diningtable 1. \\
\textcolor{thought}{Thought}: Now I take a lettuce 1. Next, I need to go to sinkbasin 1 and clean it.\\ 
\textcolor{action}{Action}: go to sinkbasin 1\\
\textcolor{obs}{Observation}: On the sinkbasin 1, you see a apple 2, a ladle 2, a spoon 1, and a tomato 3. \\
\textcolor{action}{Action}: clean lettuce 1 with sinkbasin 1\\
\textcolor{obs}{Observation}: You clean the lettuce 1 using the sinkbasin 1.\\
\textcolor{thought}{Thought}: Now I clean a lettuce 1. Next, I need to put it in/on diningtable 1.\\ 
\textcolor{action}{Action}: go to diningtable 1\\
\textcolor{obs}{Observation}: On the diningtable 1, you see a apple 1, a bread 1, a butterknife 2, a cup 2, a fork 2, a knife 2, a knife 1, a ladle 1, a mug 2, a mug 1, a pan 2, a peppershaker 1, a spatula 3, a tomato 2, and a tomato 1.\\
\textcolor{action}{Action}: put lettuce 1 in/on diningtable 1 \\
\textcolor{obs}{Observation}: You put the lettuce 1 in/on the diningtable 1.
\end{tcolorbox}
\captionof{figure}{Prompt for ALFWorld ReAct, "pick clean then place" task}

\vspace{1cm}

\noindent
\centering
\begin{tcolorbox}[
  colback=white,
  colframe=gray,
  enhanced,
  fontupper=\small\fontfamily{pcr}\selectfont,
  sharp corners=southwest,
  coltitle=black,
  colbacktitle=gray!10,
  title={\small{Prompt for ALFWorld ReflAct, "pick clean then place" task}},
  breakable
]

You are in the middle of a room. Looking quickly around you, you see a cabinet 13, a cabinet 12, a cabinet 11, a cabinet 10, a cabinet 9, a cabinet 8, a cabinet 7, a cabinet 6, a cabinet 5, a cabinet 4, a cabinet 3, a cabinet 2, a cabinet 1, a coffeemachine 1, a countertop 1, a diningtable 1, a drawer 1, a fridge 1, a garbagecan 1, a microwave 1, a shelf 3, a shelf 2, a shelf 1, a sinkbasin 1, a stoveburner 4, a stoveburner 3, a stoveburner 2, a stoveburner 1, and a toaster 1.\\
Your task is to: put a clean lettuce in diningtable. \\\\
\textcolor{reflection}{Reflection}: To solve the task, I need to find and take a lettuce, then clean it with sinkbasin, then put it in diningtable. First I need to find a lettuce. A lettuce is more likely to appear in fridge 1, diningtable 1, sinkbasin 1, stoveburner 1-3, cabinet 1-13. I can check one by one, starting by going to fridge 1.\\
\textcolor{action}{Action}: go to fridge 1 \\
\textcolor{obs}{Observation}: The fridge 1 is closed.\\
\textcolor{reflection}{Reflection}: Currently, I am in front of the closed fridge 1, not holding anything, and need to open it to search for lettuce and continue the task.\\
\textcolor{action}{Action}: open fridge 1\\
\textcolor{obs}{Observation}: You open the fridge 1. The fridge 1 is open. In it, you see a cup 3, a egg 2, a potato 3, and a potato 2.\\
\textcolor{reflection}{Reflection}: Currently, I am in front of open fridge 1, not holding any object, and still need to find the lettuce to complete the task.\\
\textcolor{action}{Action}: go to diningtable 1\\
\textcolor{obs}{Observation}: On the diningtable 1, you see a apple 1, a bread 1, a butterknife 2, a cup 2, a fork 2, a knife 2, a knife 1, a ladle 1, a lettuce 1, a mug 2, a mug 1, a pan 2, a peppershaker 1, a spatula 3, a tomato 2, and a tomato 1. \\
\textcolor{reflection}{Reflection}: Currently, I am at dining table 1 and have found lettuce 1, which I need to clean before placing it on the dining table to complete the task.\\
\textcolor{action}{Action}: take lettuce 1 from diningtable 1 \\
\textcolor{obs}{Observation}: You pick up the lettuce 1 from the diningtable 1. \\
\textcolor{reflection}{Reflection}: Currently, I am at dining table 1, now holding lettuce 1, which I need to clean before placing it on the dining table to complete the task. \\
\textcolor{action}{Action}: go to sinkbasin 1\\
\textcolor{obs}{Observation}: On the sinkbasin 1, you see a apple 2, a ladle 2, a spoon 1, and a tomato 3. \\
\textcolor{reflection}{Reflection}: Currently, I am at sinkbasin 1, I’m holding a lettuce 1, and I am now ready to clean it before placing it on the dining table to complete my task.\\
\textcolor{action}{Action}: clean lettuce 1 with sinkbasin 1\\
\textcolor{obs}{Observation}: You clean the lettuce 1 using the sinkbasin 1.\\ 
\textcolor{reflection}{Reflection}: Currently, I am at sinkbasin 1 and I am holding a clean lettuce 1, while the task of putting the clean lettuce on the dining table is almost complete as I just need to return to the dining table.\\
\textcolor{action}{Action}: go to diningtable 1\\
\textcolor{obs}{Observation}: On the diningtable 1, you see a apple 1, a bread 1, a butterknife 2, a cup 2, a fork 2, a knife 2, a knife 1, a ladle 1, a mug 2, a mug 1, a pan 2, a peppershaker 1, a spatula 3, a tomato 2, and a tomato 1.\\
\textcolor{reflection}{Reflection}: Currently, I am at diningtable 1 and I am holding a clean lettuce 1, while the task of putting the clean lettuce on the dining table is now complete as I just need to place it down.\\
\textcolor{action}{Action}: put lettuce 1 in/on diningtable 1 \\
\textcolor{obs}{Observation}: You put the lettuce 1 in/on the diningtable 1.
\end{tcolorbox}
\captionof{figure}{Prompt for ALFWorld ReflAct, "pick clean then place" task}

\newpage

\subsection{ScienceWorld}
\subsubsection{Instructions}
\vspace{0.5cm}
\noindent
\centering
\begin{tcolorbox}[
  colback=white,
  colframe=gray,
  enhanced,
  fontupper=\small\fontfamily{pcr}\selectfont,
  sharp corners=southwest,
  coltitle=black,
  colbacktitle=gray!10,
  title={\small{Instruction for ScienceWorld Nothinking, ReAct, ReflAct}}
]
You are a helpful assistant to do some scientific experiment in an environment.
In the environment, there are several rooms: kitchen, foundry, workshop, bathroom, outside, living room, bedroom, greenhouse, art studio, hallway\\
You should explore the environment and find the items you need to complete the experiment.\\
You can teleport to any room in one step.\\
All containers in the environment have already been opened, you can directly get items from the containers.\\
For each of your turn, you will be given the observation of the last turn. \\
\textcolor{nothinking}{(Nothinking)} You should directly output the action in this turn. Your output must strictly follow this format:"Action: your next action". \\
\textcolor{thought}{(ReAct)} You should first think about the current condition and plan for your future actions, and then output your action in this turn. Your output must strictly follow this format:"Thought: your thoughts.\textbackslash n Action: your next action". Remember that you can only output one "Action:" in per response.\\
\textcolor{reflection}{(ReflAct)} You should first reflect on the agent's state, including the location, inventory, and focused object, in relation to the task goal. Then, output the action for this turn. Your output must strictly follow this format:"Reflection: your reflection.\textbackslash n Action: your next action".\\

The available actions are:\\
open OBJ: open a container\\
close OBJ: close a container\\
activate OBJ: activate a device\\
deactivate OBJ: deactivate a device\\
connect OBJ to OBJ: connect electrical components\\
disconnect OBJ: disconnect electrical components\\
use OBJ [on OBJ]: use a device/item\\
look around: describe the current room\\
examine OBJ: describe an object in detail\\
look at OBJ: describe a container's contents\\
read OBJ: read a note or book\\
move OBJ to OBJ: move an object to a container\\
pick up OBJ: move an object to the inventory\\
pour OBJ into OBJ: pour a liquid into a container\\
mix OBJ: chemically mix a container\\
teleport to LOC: teleport to a specific room\\
focus on OBJ: signal intent on a task object\\
wait: task no action for 10 steps\\
wait1: task no action for a step
\end{tcolorbox}
\captionof{figure}{Instruction for ScienceWorld Nothinking, ReAct, ReflAct}

\newpage
\subsubsection{ICL Examples}

\vspace{0.5cm}

\begin{tcolorbox}[
  colback=white,
  colframe=gray,
  enhanced,
  fontupper=\small\fontfamily{pcr}\selectfont,
  coltitle=black,
  colbacktitle=gray!10,
  title={\small{Prompt for ScienceWorld ReAct}}
]

Your task is to use chemistry to create green paint. When you are done, focus on the green paint.

\textcolor{thought}{Thought}: To solve the task, I need to find materials to create green paint, use those materials, and focus on the object specified in the task once it is identified. The materials for making green paint are likely to be found in an art studio.\\\textcolor{action}{Action}: look around

\textcolor{obs}{Observation}: This room is called the bedroom. In it, you see: \\ \qquad the agent\\ \qquad substance called air\\ \qquad bed. On the bed is: a mattress. On the mattress is: a white pillow..\\ \qquad closet. The closet door is open. In the closet is: nothing.\\ \qquad picture\\ \qquad table. On the table is: nothing.\\You also see:\\ \qquad A door to the hallway (that is open)

\textcolor{thought}{Thought}: The bedroom doesn't have the materials I need. I should go to a place where paint and painting supplies are likely to be found, such as an art studio.\\\textcolor{action}{Action}: teleport to art studio

\textcolor{obs}{Observation}: You teleport to the art studio.

\textcolor{thought}{Thought}: Now that I'm in the art studio, I should look around to find the necessary materials to create green paint.\\\textcolor{action}{Action}: look around

\textcolor{obs}{Observation}: This room is called the art studio. In it, you see: \\\qquad the agent\\ \qquad substance called air\\\qquad large cupboard. The large cupboard door is open. In the large cupboard is: a drawer, a wood cup (containing yellow paint), a wood cup (containing blue paint), a wood cup (containing red paint).\\\qquad table. On the table is: a glass cup (containing nothing).\\\qquad wood cup (containing yellow paint)\\ \qquad a wood cup (containing blue paint)\\ \qquad wood cup (containing red paint)\\You also see:\\\qquad A door to the hallway (that is open)

\textcolor{thought}{Thought}: I see blue paint here, which is one of the primary colors needed to mix green paint. I should start by pouring it into an empty container to mix with another color.\\\textcolor{action}{Action}: pour cup containing blue paint in art studio in cup containing nothing

\textcolor{obs}{Observation}: You pour the contents of the wood cup into the glass cup.

\textcolor{thought}{Thought}: Yellow paint is the other primary color needed to mix with blue to create green. I should add the yellow paint to the container with blue paint.\\\textcolor{action}{Action}: pour cup containing yellow paint in art studio in cup containing blue paint in table

\textcolor{obs}{Observation}: You pour the contents of the wood cup into the glass cup.

\textcolor{thought}{Thought}: Now that I have both blue and yellow paint in the same container, I need to mix them together to create green paint.\\\textcolor{action}{Action}: mix cup containing blue paint and yellow paint

\textcolor{obs}{Observation}: Blue and yellow paint mix to produce green paint.

\textcolor{thought}{Thought}: The task is to create green paint and then focus on it. Now that the green paint is made, I should focus on the final product.\\\textcolor{action}{Action}: focus on green paint

\end{tcolorbox}
\captionof{figure}{Prompt for ScienceWorld ReAct}

\newpage

\begin{tcolorbox}[
  colback=white,
  colframe=gray,
  enhanced,
  fontupper=\small\fontfamily{pcr}\selectfont,
  coltitle=black,
  colbacktitle=gray!10,
  title={\small{Prompt for ScienceWorld ReflAct}}
]

Your task is to use chemistry to create green paint. When you are done, focus on the green paint.

\textcolor{reflection}{Reflection}: To solve the task, I need to find materials to create green paint, use those materials, and focus on the object specified in the task once it is identified. The materials for making green paint are likely to be found in an art studio.\\\textcolor{action}{Action}: look around

\textcolor{obs}{Observation}: This room is called the bedroom. In it, you see: \\ \qquad the agent\\ \qquad substance called air\\ \qquad bed. On the bed is: a mattress. On the mattress is: a white pillow..\\ \qquad closet. The closet door is open. In the closet is: nothing.\\ \qquad picture\\ \qquad table. On the table is: nothing.\\You also see:\\ \qquad A door to the hallway (that is open)

\textcolor{reflection}{Reflection}: Currently, I am in the bedroom, holding nothing and focused on nothing. There are no useful materials here to create the green paint I need to focus on.\\\textcolor{action}{Action}: teleport to art studio

\textcolor{obs}{Observation}: You teleport to the art studio.

\textcolor{reflection}{Reflection}: Currently, I am in the art studio, holding nothing and focused on nothing. I need to look for materials to create green paint.\\\textcolor{action}{Action}: look around

\textcolor{obs}{Observation}: This room is called the art studio. In it, you see: \\\qquad the agent\\ \qquad substance called air\\\qquad large cupboard. The large cupboard door is open. In the large cupboard is: a drawer, a wood cup (containing yellow paint), a wood cup (containing blue paint), a wood cup (containing red paint).\\\qquad table. On the table is: a glass cup (containing nothing).\\\qquad wood cup (containing yellow paint)\\ \qquad a wood cup (containing blue paint)\\ \qquad wood cup (containing red paint)\\You also see:\\\qquad A door to the hallway (that is open)

\textcolor{reflection}{Reflection}: Currently, I am in the art studio, not holding anything. I’ve found the primary colors yellow, blue, and red, which can be used to make green paint, the substance I will need to focus on.\\\textcolor{action}{Action}: pour cup containing blue paint in art studio in cup containing nothing

\textcolor{obs}{Observation}: You pour the contents of the wood cup into the glass cup.

\textcolor{reflection}{Reflection}: I poured blue paint into the glass cup. I’m now holding it and will next add yellow paint.\\\textcolor{action}{Action}: pour cup containing yellow paint in art studio in cup containing blue paint in table

\textcolor{obs}{Observation}: You pour the contents of the wood cup into the glass cup.

\textcolor{reflection}{Reflection}: I’ve added yellow paint to the blue in the glass cup. I’m now holding a mix of blue and yellow paint, ready to be mixed into green.\\\textcolor{action}{Action}: mix cup containing blue paint and yellow paint

\textcolor{obs}{Observation}: Blue and yellow paint mix to produce green paint.

\textcolor{reflection}{Reflection}: The task is to create green paint and then focus on it. Now that the green paint is made, I should focus on the final product.\\\textcolor{action}{Action}: focus on green paint

\end{tcolorbox}
\captionof{figure}{Prompt for ScienceWorld ReflAct}

\newpage
\subsection{Jericho}
\subsubsection{Instructions}
\begin{figure*}[h!]
\noindent
\centering
\begin{tcolorbox}[
  colback=white,
  colframe=gray,
  enhanced,
  fontupper=\small\fontfamily{pcr}\selectfont,
  sharp corners=southwest,
  coltitle=black,
  colbacktitle=gray!10,
  title={\small{Instruction for Jericho Nothinking, ReAct, ReflAct}}
]

\textcolor{nothinking}{(Nothinking)} You are a game master in fictional text games.\\

\textcolor{thought}{(ReAct)} You are a game master in fictional text games. For each of your turn, you will be given the observation of the last turn. You should first think about the current condition and plan for your future actions, and then output your action for this turn. Your output must strictly follow this format:\"Thought: your thoughts.\textbackslash n Action: your next action\".\\

\textcolor{reflection}{(ReflAct)} You are a game master in fictional text games. For each of your turn, you will be given the observation of the last turn. You should first reflect on the agent’s state in relation to the task goal, and then output the action for this turn. Your output must strictly follow this format:``Reflection: your reflection. \textbackslash n Action: your next action''.\\

You are in a fictional game environment and you need to accomplish goals by performing actions. Each action is a simple phrase. Here are the actions you can do: \\
Inventory: check things you are carrying\\
Look: check your surroundings\\
Examine <place/obj>: check the details of something\\
Take <obj>: pickup obj\\ 
Put down <obj>: leave a obj at your current place.\\
Drop <obj>\\
Check valid actions: Check actions you can use\\
South: go south\\
North: go north\\
East: go east\\
West: go west\\
Up: go up\\
Down: go down\\
Other available actions could be determined through check valid actions.
\end{tcolorbox}
\captionof{figure}{Instruction for Jericho Nothinking, ReAct, ReflAct}
\end{figure*}

\subsubsection{ICL Examples}
\vspace{0.5cm}

\begin{tcolorbox}[
  colback=white,
  colframe=gray,
  enhanced,
  fontupper=\small\fontfamily{pcr}\selectfont,
  title={\small{Prompt for Jericho ReAct}},
  coltitle=black,
  colbacktitle=gray!10,
  breakable
]
Goal: You are the warrior Link that needs to save the princess from the castle.

\textcolor{obs}{Observation}: You are at the path leading to the castle. The castle is to your north. There is a barrel in front of you. 

\textcolor{thought}{Thought}: To solve the task, I first need to explore my surroundings to gather any useful items. Once prepared, I will move north into the castle, navigate to the princess’s location, interact with the princess and complete the rescue mission.

\textcolor{action}{Action}: Examine barrel

\textcolor{obs}{Observation}: There is a gleaming elvish sword. 

\textcolor{thought}{Thought}: The sword inside the barrel has a gleaming finish, which suggests it is of high quality and possibly enchanted. Taking this sword now will likely increase my chances of handling any immediate challenges.

\textcolor{action}{Action}: take the sword

\textcolor{obs}{Observation}: The sword is taken

\textcolor{thought}{Thought}: I'm now armed. Advancing north into the castle should provide new opportunities for exploration and reveal the next critical steps.

\textcolor{action}{Action}: north

\textcolor{obs}{Observation}:  Main hall \\ The castle is dimly lit, with lava flowing in many places. There is a staircase leading up to princess bedroom, a door to your north leading to the kitchen hall, and the door you just entered on your south

\textcolor{thought}{Thought}: Before making a move toward any specific direction, I need to evaluate what options are actually available to me right now. It would be helpful to check the possible actions I can take from here.

\textcolor{action}{Action}: check valid actions

\textcolor{obs}{Observation}: south, north, up, look, examine staircase

\textcolor{thought}{Thought}: Among the available actions, ascending the staircase seems like the most direct route to explore. Since it leads upward and the hall below seems perilous, I’ll proceed to investigate what’s at the top of the stairs.

\textcolor{action}{Action}: up

\textcolor{obs}{Observation}: Princess’s bedroom. Princess is sleeping in bed. 

\textcolor{thought}{Thought}: I’ve entered a bedroom where the princess is lying unconscious. There are likely specific actions I can perform in this situation, and I should understand my options before taking a step that could have consequences. Checking what I can do here would be the most logical next move.

\textcolor{action}{Action}: check valid actions

\textcolor{obs}{Observation}: wake up the princess,  take out sword, down

\textcolor{thought}{Thought}: One of the available actions is to wake the princess, which might change the course of events. It seems like the most immediate and gentle way to interact with her. I’ll go ahead and try that first.

\textcolor{action}{Action}: wake up the princess

\textcolor{obs}{Observation}: The princess wake up from the coma. Thank you my knight, she says. The task is finished.

\end{tcolorbox}
\captionof{figure}{Prompt for Jericho ReAct}



\vspace{1cm}

\begin{tcolorbox}[
  colback=white,
  colframe=gray,
  enhanced,
  fontupper=\small\fontfamily{pcr}\selectfont,
  coltitle=black,
  colbacktitle=gray!10,
  title={\small{Prompt for Jericho ReflAct}},
  breakable,
]

Goal: You are the warrior Link that needs to save the princess from the castle.

\textcolor{obs}{Observation}: You are at the path leading to the castle. The castle is to your north. There is a barrel in front of you. 

\textcolor{reflection}{Reflection}: To solve the task, I first need to explore my surroundings to gather any useful items. Once prepared, I will move north into the castle, navigate to the princess’s location, interact with the princess and complete the rescue mission. 

\textcolor{action}{Action}: Examine barrel

\textcolor{obs}{Observation}: There is a gleaming elvish sword. 

\textcolor{reflection}{Reflection}: I found a powerful elvish sword in the barrel. Since I currently have no weapons, this sword will be essential for protection once I enter the castle.

\textcolor{action}{Action}: take the sword

\textcolor{obs}{Observation}: The sword is taken

\textcolor{reflection}{Reflection}: I have obtained the elvish sword and am now equipped. With my weapon ready, I should head into the castle to continue the search for the princess.

\textcolor{action}{Action}: north

\textcolor{obs}{Observation}: Main hall \\ The castle is dimly lit, with lava flowing in many places. There is a staircase leading up to princess bedroom, a door to your north leading to the kitchen hall, and the door you just entered on your south

\textcolor{reflection}{Reflection}: I have entered the main hall of the castle. The environment is dangerous with lava present. The staircase likely leads to the princess’s bedroom, which aligns with my goal. I should verify available actions first.

\textcolor{action}{Action}: check valid actions

\textcolor{obs}{Observation}: south, north, up, look, examine staircase

\textcolor{reflection}{Reflection}: The valid actions include going “up,” which matches the staircase that leads to the princess's bedroom. This is likely the shortest path to the goal.

\textcolor{action}{Action}: up

\textcolor{obs}{Observation}: Princess’s bedroom. Princess is sleeping in bed.

\textcolor{reflection}{Reflection}: I have reached the princess, but she appears to be unconscious or in a coma. I need to determine how to safely wake her to complete the mission.

\textcolor{action}{Action}: check valid actions

\textcolor{obs}{Observation}: wake up the princess,  take out sword, down

\textcolor{reflection}{Reflection}: The available actions include “wake up the princess,” which directly supports the goal of saving her. No threat is present, so this is the appropriate move.

\textcolor{action}{Action}: wake up the princess

\textcolor{obs}{Observation}: The princess wake up from the coma. Thank you my knight, she says. The task is finished.
\end{tcolorbox}
\captionof{figure}{Prompt for Jericho ReflAct}

\newpage
\section{More Case Study}

\begin{figure*}[h!]
\begin{tcolorbox}[
  enhanced,
  title=\small{\textbf{ALFWorld}: clean some soapbar and put it in cabinet.},
  colback=white,
  coltitle=black, 
  colframe=gray,
  colbacktitle=gray!10, 
  fontupper=\footnotesize\fontfamily{pcr}\selectfont,
  fontlower=\footnotesize\fontfamily{pcr}\selectfont,
  sidebyside,
  sidebyside gap=5mm,
  sidebyside align=top,
  width=\textwidth,
  enhanced
]
\footnotesize
\textbf{ReAct}

Goal: clean some soapbar and put it in cabinet.\\
\textcolor{thought}{Thought}: To solve the task, I need to find a soapbar, then clean it with sinkbasin, then put it in cabinet. First I need to find a soapbar. A soapbar is more likely to appear in cabinet 1-4, countertop 1, sinkbasin 1-2, or toilet 1. I can check one by one, starting by going to cabinet 1.\\
\textcolor{action}{Action}: go to cabinet 1\\
...\\
\textcolor{action}{Action}: go to cabinet 4\\
\textcolor{obs}{Observation}: You open the cabinet 4. The cabinet 4 is open. In it, you see a candle 1, and a cloth 2.\\
\textcolor{thought}{Thought}: Now I find a cloth 2 in cabinet 4, which I can use to clean the soapbar. Next, I need to find a soapbar. I will go to countertop 1 to check if it contains a soapbar.\\
\textcolor{action}{Action}: go to countertop 1\\
\textcolor{obs}{Observation}: On the countertop 1, you see a soapbar 1, and a soapbottle 1.\\
\textcolor{thought}{Thought}: \textbf{Now I find a soapbar 1 on the countertop 1. Next, I need to clean it with sinkbasin. I will go to sinkbasin 1 to clean the soapbar.}\textcolor{thought}{$^*$}\\
\textcolor{thought}{\quad $^*$ Ungrounded thought: The agent attempts to go to the sinkbasin without having taken the soapbar.}\\
\\
\textcolor{action}{Action}: go to sinkbasin 1\\
\textcolor{obs}{Observation}: On the sinkbasin 1, you see nothing.\\
\textcolor{thought}{Thought}: Since the sinkbasin 1 is empty, I need to clean the soapbar 1 with something. \textbf{I will use the cloth 2 from cabinet 4 to clean the soapbar.\textcolor{thought}{$^*$}} \\
\textcolor{thought}{\quad $^*$ Hallucination: The agent performs an action without soapbar present.}\\
\\
\textcolor{action}{Action}: \textbf{use cloth 2}\\
\textcolor{obs}{Observation}: Nothing happens.\\
\textcolor{thought}{Thought}: Since using the cloth 2 did not have any effect, I need to find a way to use it to clean the soapbar. I will go back to the cabinet 4 and take the cloth 2 to use it with the sinkbasin 1.\\
\textcolor{action}{Action}: take cloth 2 from cabinet 4\\
\textcolor{obs}{Observation}: Nothing happens.\\
...\\
\textcolor{red}{\textbf{\# Task Failed}}

\tcblower
\footnotesize
\textbf{ReflAct}

Goal: clean some soapbar and put it in cabinet.\\
\textcolor{reflection}{Reflection}: To solve the task, I need to find a soapbar, clean it with sinkbasin, and then put it in cabinet. First, I need to find a soapbar. A soapbar is more likely to appear in cabinet 1-4, countertop 1, or sinkbasin 1-2. I can check one by one, starting by going to cabinet 1.\\
\textcolor{action}{Action}: go to cabinet 1\\
...\\
\textcolor{action}{Action}: go to countertop 1\\
\textcolor{obs}{Observation}: On the countertop 1, you see a soapbar 1, and a soapbottle 1.\\
\textcolor{reflection}{Reflection}: Currently, I am at countertop 1, and \textbf{I have found a soapbar 1, which I need to clean before placing it in the cabinet to complete the task.\textcolor{reflection}{$^*$}}\\
\textcolor{reflection}{\quad $^*$ The agent performs an action that are goal-related and well grounded in the current situation.}\\
\\
\textcolor{action}{Action}: \textbf{take soapbar 1 from countertop 1}\\
\textcolor{obs}{Observation}: You pick up the soapbar 1 from the countertop 1.\\
\textcolor{reflection}{Reflection}: Currently, I am at countertop 1, now holding soapbar 1, which I need to clean before placing it in the cabinet to complete my task.\\
\textcolor{obs}{Observation}: On the sinkbasin 1, you see nothing.\\
\textcolor{reflection}{Reflection}:Currently, I am at sinkbasin 1, I'm holding a soapbar 1, and I am now ready to clean it before placing it in the cabinet to complete my task.\\
\textcolor{action}{Action}: Action: clean soapbar 1 with sinkbasin 1\\
\textcolor{obs}{Observation}: You clean the soapbar 1 using the sinkbasin 1.\\
...\\
\textcolor{custom_green}{\textbf{\# Task Succeed}}

\end{tcolorbox}
\caption{Case study for ALFWorld}
\end{figure*}

\newpage
\begin{figure*}[h!]
\begin{tcolorbox}[
  enhanced,
  title=\small{\textbf{ALFWorld}: examine the book with the desklamp.},
  colback=white,
  coltitle=black, 
  colframe=gray,
  colbacktitle=gray!10, 
  fontupper=\footnotesize\fontfamily{pcr}\selectfont,
  fontlower=\footnotesize\fontfamily{pcr}\selectfont,
  sidebyside,
  sidebyside gap=5mm,
  sidebyside align=top,
  width=\textwidth,
  enhanced
]
\footnotesize
\textbf{ReAct}

Goal: examine the book with the desklamp.\\
\textcolor{thought}{Thought}: To accomplish the task, I need to find a book and a desklamp. The book is likely to be found on the desk or shelves, while the desklamp is likely to be on one of the desks or shelves. I will start by checking desk 1 for a book.\\
\textcolor{action}{Action}: go to desk 1\\
\textcolor{obs}{Observation}: On the desk 1, you see a alarmclock 2, a creditcard 1, a desklamp 1, and a keychain 2.\\
...\\
\textcolor{obs}{Observation}: Nothing happens.\\
\textcolor{thought}{Thought}: The desklamp is not functioning as expected, and I have exhausted all options for finding a book in the environment. Since I cannot complete the task of examining a book with the desklamp, \textbf{I will return to desk 1 to reassess my options and see if there is anything I might have missed.}\textcolor{thought}{$^*$}\\
\textcolor{thought}{\quad $^*$ Unnecessary exploitation: the agent takes a redundant action by revisiting desk 1, which it has already explored.}\\
\textcolor{action}{Action}: go to desk 1\\
\textcolor{obs}{Observation}: On the desk 1, you see a alarmclock 2, a creditcard 1, a desklamp 1, and a keychain 2.\\
\textcolor{thought}{Thought}: I see the same items on desk 1 again, confirming that there is no book here. Since I have checked all possible locations and have not found a book, I will now check the garbage can, as sometimes items can be discarded there. 
 \\
\textcolor{action}{Action}: go to garbagecan 1 \\
\textcolor{obs}{Observation}: On the garbagecan 1, you see nothing.\\
\textcolor{thought}{Thought}: There is nothing in the garbage can. I have searched all the desks, shelves, drawers, the laundry hamper, and the garbage can without finding a book. It seems I have exhausted all possible locations for a book in this environment. Since I cannot complete the task, \textbf{I will attempt to use the desklamp again in case it can provide any assistance or reveal something I missed.}\textcolor{thought}{$^*$}\\
\textcolor{thought}{\quad $^*$ Ungrounded thought: The agent incorrectly attempts to use the desklamp on desk 1, which is not at its current location.}\\
\textcolor{action}{Action}: use desklamp 1 \\
\textcolor{obs}{Observation}: Nothing happens.\\
...\\
\textcolor{red}{\textbf{\# Task Failed}}

\tcblower
\footnotesize
\textbf{ReflAct}

Goal: examine the book with the desklamp.\\
\textcolor{reflection}{Reflection}: To accomplish my task of examining the book with the desklamp, I first need to locate a book and then find the desklamp to use it for illumination. I will start by checking drawer 1 for a book. \\
\textcolor{action}{Action}: go to drawer 1\\
\textcolor{obs}{Observation}: The drawer 1 is closed.\\
...\\
\textcolor{reflection}{Reflection}: Currently, I am at drawer 6, and I am not holding any object, as my search for a book has not been successful, which is preventing me from completing my task of examining it with the desklamp.  \\
\textcolor{action}{Action}: go to desk 1 \\
\textcolor{obs}{Observation}: On the desk 1, you see a alarmclock 1, a creditcard 1, a desklamp 1, a laptop 3, and a mug 1.\\
...\\
\textcolor{reflection}{Reflection}: Currently, I am at shelf 6, and I am not holding any object, \textbf{as my search for a book has not yielded any results, which is preventing me from completing my task of examining it with the desklamp.}\textcolor{reflection}{$^*$} \\
\textcolor{reflection}{\quad $^*$ The agent performs an action that are goal-related and well grounded in the current situation.}\\
\\
\textcolor{action}{Action}: go to bed 1
\\
\textcolor{obs}{Observation}: On the bed 1, you see a book 1, a cellphone 1, a laptop 2, a laptop 1, a pillow 2, and a pillow 1.\\
\textcolor{reflection}{Reflection}: Currently, I am at bed 1, and I am not holding any object, \textbf{but I have finally located a book, which is essential for completing my task of examining it with the desklamp.} \\

\textcolor{action}{Action}: take book 1 from bed 1\\

\textcolor{obs}{Observation}: You pick up the book 1 from the bed 1. \\
\textcolor{reflection}{Reflection}: Currently, I am at bed 1, and I am holding the book 1 I just picked up, which allows me to proceed with my task of examining it with the desklamp. \\
\textcolor{action}{Action}: go to desk 1 \\
\textcolor{obs}{Observation}: On the desk 1, you see a alarmclock 1, a creditcard 1, a desklamp 1, a laptop 3, and a mug 1.\\
\textcolor{reflection}{Reflection}: Currently, I am at desk 1, holding the book 1 I picked up from bed 1, and I can now proceed to examine it with the desklamp that is also located on this desk.  \\
\textcolor{action}{Action}: use desklamp 1 \\
...\\
\textcolor{custom_green}{\textbf{\# Task Succeed}}

\end{tcolorbox}
\caption{Case study for ALFWorld}
\end{figure*}

\newpage

\begin{figure*}[!t]
\vspace{-5mm}
\begin{tcolorbox}[
  enhanced,
  title={\shortstack[l]{\small{\textbf{ScienceWorld}: Your task is to boil tin.} \\
\small{For compounds without a boiling point, combusting the substance is also acceptable.} \\
\small{First, focus on the substance. Then, take actions that will cause it to change its state of matter.}}},
  colback=white,
  coltitle=black, 
  colframe=gray,
  colbacktitle=gray!10, 
  fontupper=\footnotesize\fontfamily{pcr}\selectfont,
  fontlower=\footnotesize\fontfamily{pcr}\selectfont,
  sidebyside,
  sidebyside gap=5mm,
  sidebyside align=top,
  width=\textwidth,
  enhanced,
]
\footnotesize
\textbf{ReAct}

...\\
\textcolor{action}{Action}: teleport to workshop\\
\textcolor{obs}{Observation}: You teleport to the workshop.\\
\textcolor{thought}{Thought}: Now that I'm in the workshop, I should look around to find tin. The workshop is a place where various projects and repairs are done, so I might find tin here in the form of scraps, sheets, or other objects.\\
\textcolor{action}{Action}: look around\\
\textcolor{obs}{Observation}: This room is called the workshop. In it, you see: ... On the table is: a battery, a green wire, a metal pot (containing a substance called tin), ...\\
\textcolor{thought}{Thought}: I found tin in the metal pot on the table! Now that I've located the tin, I should focus on it as per the task instructions.\\
\textcolor{action}{Action}: focus on tin\\
\textcolor{obs}{Observation}: You focus on the tin.\\
\textcolor{thought}{Thought}: Now that I've focused on the tin, I need to take actions to boil or combust it. \textbf{Since the tin is in a metal pot, I can use a heat source to boil it. I see a blast furnace in the foundry, which would be perfect for heating the tin. I should teleport back to the foundry and use the blast furnace.\textcolor{thought}{$^*$}}\\
\textcolor{thought}{\quad $^*$ Ungrounded thought: The agent confirms the presence of tin and proceeds with planning, but does not actually possess the tin.}\\
\\
\textcolor{action}{Action}: teleport to foundry\\
\textcolor{obs}{Observation}: You teleport to the foundry.\\
\textcolor{thought}{Thought}: Now that I'm back in the foundry, I should move the metal pot containing the tin to the blast furnace and then activate the blast furnace to heat the tin.\\
\textcolor{action}{Action}: move metal pot to blast furnace\\
\textcolor{obs}{Observation}: No known action matches that input.\\
\textcolor{thought}{Thought}: \textbf{It seems I made a mistake with the action. I should try again, using a different action to move the metal pot to the blast furnace.\textcolor{thought}{$^*$}}\\
\textcolor{thought}{\quad $^*$ Hallucination: 
Although the agent recognizes the mistake, it continues to attempt the same invalid action.}\\
\\
\textcolor{action}{Action}: \textbf{move metal pot to blast furnace}\\
...\\
\textcolor{red}{\textbf{\# Task Failed}}
\tcblower
\footnotesize
\textbf{ReflAct}

...\\
\textcolor{action}{Action}: teleport to workshop\\
\textcolor{obs}{Observation}: You teleport to the workshop.\\
\textcolor{reflection}{Reflection}: Currently, I am in the workshop, holding nothing and focused on nothing. I need to look for tin in this room.\\
\textcolor{action}{Action}: look around\\
\textcolor{obs}{Observation}: This room is called the workshop. In it, you see: ... On the table is: a battery, a green wire, a metal pot (containing a substance called tin), ...\\
\textcolor{reflection}{Reflection}: Currently, I am in the workshop, holding nothing and focused on nothing. I've found a metal pot containing tin on the table, which is the substance I need to focus on and then boil or combust\\
\textcolor{action}{Action}: focus on tin\\
\textcolor{obs}{Observation}: You focus on the tin.\\
\textcolor{reflection}{Reflection}: I am now focused on the tin, which is contained in the metal pot on the table. To boil or combust the tin, I need to apply heat to it. I see a foundry in the environment, which is likely to have a heat source.\\
\textcolor{action}{Action}: teleport to foundry\\
...\\
\textcolor{reflection}{Reflection}: The blast furnace is now activated and ready to be used as a heat source. \textbf{I need to move the tin from the metal pot in the workshop to the blast furnace in the foundry.}\\
\textcolor{action}{Action}: teleport to workshop\\
\textcolor{obs}{Observation}: You teleport to the workshop.\\
\textcolor{reflection}{Reflection}: I am now back in the workshop, where the metal pot containing the tin is located. I need to move the tin to the blast furnace in the foundry.\\
\textcolor{action}{Action}: move metal pot to blast furnace\\
\textcolor{obs}{Observation}: No known action matches that input.\\
\textcolor{reflection}{Reflection}: \textbf{I need to move the tin to the blast furnace, but the action is not recognized. Instead, I can pick up the metal pot and then teleport to the foundry.}\\
\textcolor{action}{Action}: pick up metal pot\\
\textcolor{obs}{Observation}: You move the metal pot to the inventory.\\
\textcolor{reflection}{Reflection}: \textbf{I am now back in the foundry, with the metal pot containing the tin in my inventory. I can now use the blast furnace to boil or combust the tin.}\\
...\\
\textcolor{custom_green}{\textbf{\# Task Succeed}}
    
\end{tcolorbox}
\caption{Case study for ScienceWorld}
\end{figure*}

\newpage
\begin{figure*}[h!]
\begin{tcolorbox}[
  enhanced,
  title={\shortstack[l]{\small{\textbf{Jericho}: You are a pizza delivery man. Today you are making a new delivery on your boss's order.} \\
\small{You need to find out by yourself who to deliver.}}},
  colback=white,
  coltitle=black, 
  colframe=gray,
  colbacktitle=gray!10, 
  fontupper=\footnotesize\fontfamily{pcr}\selectfont,
  fontlower=\footnotesize\fontfamily{pcr}\selectfont,
  sidebyside,
  sidebyside gap=5mm,
  sidebyside align=top,
  width=\textwidth,
  enhanced
]
\footnotesize
\textbf{ReAct}

...\\
\textcolor{obs}{Observation}: The note reads: "Stanley, Don't forget to make your delivery to Mr. Calzone, located at the San Doppleton Courthouse. ...\\
...\\
\textcolor{obs}{Observation}: Street, Near Courthouse You are approaching the center of downtown San Doppleton. There is a courthouse to the south. The street runs east and west.\\
\textcolor{thought}{Thought}: It seems like I'm on the right track, as the courthouse is now visible to the south. I should head south to reach the courthouse and deliver the pizza to Mr. Calzone as instructed in the note.\\
\textcolor{action}{Action}: South\\
\textcolor{obs}{Observation}: Courthouse You are in a crowded courthouse. The street is off to the north. The courtroom is currently abuzz with excitement as it seems that the pre-trial hearing of the infamous gangster, Guido "The Armadillo" Calzone is winding to a close. Guido and his hotshot lawyer sit placidly at their table, while the flustered young district attorney who is prosecuting the case seems like she's at her wits' end. At the moment, the D.A. stands at the front of the courtroom, arguing over a legal point with the judge.\\
\textcolor{thought}{Thought}: I have arrived at the courthouse where Mr. Calzone is present. Since the courtroom is busy, \textbf{I need to find a way to deliver the pizza to Mr. Calzone without causing a disruption. I should approach him or his lawyer to hand over the pizza box.} \\
\textcolor{action}{Action}: \textbf{Show people box} \\ 
\textcolor{obs}{Observation}: You can't see any such thing.\\
\textcolor{thought}{Thought}: \textbf{It seems that the action "show people box" was not understood in this context. I need to find a more direct way to deliver the pizza to Mr.Calzone. Since he is present in the courtroom, I should try to give the pizza box directly to him or his lawyer.\textcolor{thought}{$^*$}}\\
\textcolor{thought}{\quad $^*$ Mismatched execution: The agent fails to specify the target, leading to another invalid action.}\\
\\
\textcolor{action}{Action}: \textbf{Give people box}\\
\textcolor{obs}{Observation}: You can't see any such thing.\\
...\\
\textcolor{red}{\textbf{\# Task Failed}}
\tcblower

\textbf{ReflAct}
\footnotesize\\
...\\
\textcolor{obs}{Observation}: The note reads: "Stanley, Don't forget to make your delivery to Mr. Calzone, located at the San Doppleton Courthouse. ...\\
...\\
\textcolor{obs}{Observation}: Street, Near Courthouse You are approaching the center of downtown San Doppleton. There is a courthouse to the south. The street runs east and west.\\
\textcolor{reflection}{Reflection}: I've finally found the courthouse, which is located to the south of my current position. This is where I need to deliver the pizza to Mr. Calzone. Now that I've located the courthouse, I should proceed south to complete the delivery task.\\
\textcolor{action}{Action}: South\\
\textcolor{obs}{Observation}: Courthouse You are in a crowded courthouse. The street is off to the north. The courtroom is currently abuzz with excitement as it seems that the pre-trial hearing of the infamous gangster, Guido "The Armadillo" Calzone is winding to a close. Guido and his hotshot lawyer sit placidly at their table, while the flustered young district attorney who is prosecuting the case seems like she's at her wits' end. At the moment, the D.A. stands at the front of the courtroom, arguing over a legal point with the judge.\\
\textcolor{reflection}{Reflection}: I've entered the courthouse where the pre-trial hearing for Guido "The Armadillo" Calzone is taking place. \textbf{This is the location where I need to deliver the pizza to Mr. Calzone. Since I've found him, I should proceed with delivering the pizza to complete my task.\textcolor{reflection}{$^*$}}\\
\textcolor{reflection}{\quad $^*$ Well-grounded and goal-aligned reasoning: The agent explicitly identifies the recipient and forms a valid, context-aware action.}\\
\\
\textcolor{action}{Action}: \textbf{Give pizza box to Guido Calzone}\\
\textcolor{obs}{Observation}: Guido grins shiftily and takes the pizza from you. He hands it to his lawyer so he can eat it later. You walk away, congratulating yourself on a job well done.\\
\textcolor{custom_green}{\textbf{\# Task Succeed}}

\end{tcolorbox}
\caption{Case study for Jericho}
\end{figure*}

\newpage
\begin{figure*}[h!]
\begin{tcolorbox}[
  enhanced,
  title=\small{\textbf{Jericho}: search for a mug in an inn.},
  colback=white,
  coltitle=black, 
  colbacktitle=gray!10, 
  colframe=gray,
  fontupper=\footnotesize\fontfamily{pcr}\selectfont,
  fontlower=\footnotesize\fontfamily{pcr}\selectfont,
  sidebyside,
  sidebyside gap=5mm,
  sidebyside align=top,
  width=\textwidth,
  enhanced,
]
\footnotesize
\textbf{ReAct}\\
...\\
\textcolor{obs}{Observation}: Inside the Inn There are a few patches of mould on the walls and the floorboards creak slightly, but the building seems safe enough. ... You see some tables and chairs. ... You can go west (back outside) or down some steps. You can see a magnificently carved chair here.\\
\textcolor{thought}{Thought}: I'm inside the inn, and I need to find a mug. I should start by examining the tables and chairs to see if there's a mug lying around.\\
\textcolor{action}{Action}: Examine tables and chairs\\
\textcolor{obs}{Observation}: You can't see any such thing.\\
...\\
\textcolor{action}{Action}: Examine floorboards\\
\textcolor{obs}{Observation}: The floor is made of large, rather uneven rough floorboards, on which lie the remains of the scattered rushes once used as a floor covering.\\
\textcolor{thought}{Thought}: I've examined the floorboards, but I didn't find a mug. I should try to find something to pick up, and since I'm in the inn, I should check the inventory again to see if I have any other options, but it seems I'm empty-handed. \textbf{I should try to find something to pick up.}\\
\textcolor{action}{Action}: Examine stone steps leading down\\
...\\
\textcolor{action}{Action}: Examine the floorboards again\\
\textcolor{obs}{Observation}: I only understood you as far as wanting to examine the floor.\\
\textcolor{thought}{Thought}: I've tried examining various objects and places in the inn, but I didn't find a mug. I should try to find something to pick up, and since I'm in the inn, I should check the inventory again to see if I have any other options, but it seems I'm empty-handed. \textbf{I should try to find something to pick up.\textcolor{thought}{$^*$}}\\
\textcolor{thought}{\quad $^*$ Short-sighted planning:
The agent repeatedly focuses narrowly on immediate surroundings without reconsidering its strategy in light of prior failures.}\\
\\
\textcolor{action}{Action}: Examine the rushes on the floor again\\
\textcolor{obs}{Observation}: I only understood you as far as wanting to examine the floor.\\
...\\
\textcolor{red}{\textbf{\# Task Failed}}
\tcblower
\footnotesize
\textbf{ReflAct}\\
...\\
\textcolor{obs}{Observation}: Inside the Inn There are a few patches of mould on the walls and the floorboards creak slightly, but the building seems safe enough. ... You see some tables and chairs. ... You can go west (back outside) or down some steps. You can see a magnificently carved chair here.\\
\textcolor{reflection}{Reflection}: I've entered the abandoned inn, the Troll's Arms, and it appears to be old and dusty, but safe enough. I've seen some tables and chairs, and a dark, intricately-carved wooden chair in the corner. \textbf{Since my goal is to search for a mug in an inn,} I should investigate the tables and chairs to see if there's anything useful.\\
\textcolor{action}{Action}: Examine tables and chairs\\
\textcolor{obs}{Observation}: You can't see any such thing.\\
...\\
\textcolor{action}{Action}: Examine bar or counter\\
\textcolor{obs}{Observation}: I only understood you as far as wanting to examine the walls.\\
\textcolor{reflection}{Reflection}: I've been trying to find a mug, but that didn't reveal anything. \textbf{I should try to think of a different approach to find the mug.}\\
\textcolor{action}{Action}: Down\\
\textcolor{obs}{Observation}: Inn Cellar The cellar seems to have been formed from a natural cavern in the rock under the inn, ... , which will remain cool in the slightly chill air. ... You can go up (to the inn) or west (into the dark tunnel). You can see an ancient silver coin and some barrels here.\\
\textcolor{reflection}{Reflection}: I've been trying to find a mug, but so far, I haven't found anything useful. \textbf{I should try to think of a different approach to find the mug. The cellar seems like a promising location to search for a mug, as it's a storage area for the inn's supplies.\textcolor{reflection}{$^*$}}\\
\textcolor{reflection}{\quad $^*$ Reflecting on long-term goal: The agent adapts its exploration strategy based on environmental cues and prior search failures.}\\
\\
\textcolor{action}{Action}: Examine barrels\\
\textcolor{obs}{Observation}: The barrels are now no more than rotten staves of dry wood, held together by the woodworm holding hands. You have found a pewter mug.\\
\textcolor{custom_green}{\textbf{\# Task Succeed}}

\end{tcolorbox}
\caption{Another case study for Jericho}
\end{figure*}

\end{document}